\documentclass[information,article,accept,pdftex,moreauthors]{Definitions/mdpi} 

\usepackage[normalem]{ulem} 

\firstpage{1} 
\pubvolume{16}
\issuenum{9}
\articlenumber{719}
\pubyear{2025}
\copyrightyear{2025}
\externaleditor{Marco Leo} 
\datereceived{3 June 2025} 
\daterevised{13 July 2025} 
\dateaccepted{12 August 2025} 
\datepublished{22 August 2025} 
\hreflink{https://doi.org/10.3390/\\info16090719} 

\usepackage{caption}
\usepackage {subfig}

\Title{Combine Virtual Reality and Machine-Learning to Identify the Presence of Dyslexia: A Cross-Linguistic Approach}

\TitleCitation{Combine Virtual Reality and Machine-Learning to Identify the Presence of Dyslexia: A Cross-\linebreak Linguistic Approach}

\Author{{Michele} Materazzini $^{1,}$*\orcidA{}, Gianluca Morciano $^{1}$, José Manuel Alcalde-Llergo $^{1}$\orcidB{}, Enrique Yeguas-Bolívar $^{2}$\orcidC{},\linebreak   Giuseppe Calabrò $^{1}$\orcidD{}, Andrea Zingoni $^{1}$\orcidE{} and Juri Taborri $^{1,}$*\orcidF{}}

\AuthorNames{Michele Materazzini, Gianluca  Morciano, José Manuel Alcalde-Llergo, Enrique Yeguas-Bolívar, Giuseppe Calabrò, Andrea Zingoni and Juri Taborri}

\isAPAStyle{%
       \AuthorCitation{Lastname, F., Lastname, F., \& Lastname, F.}
         }{%
        \isChicagoStyle{%
        \AuthorCitation{Lastname, Firstname, Firstname Lastname and Firstname Lastname.}
        }{
        \AuthorCitation{Materazzini, M.; Morciano, G.; Alcalde-Llergo, J.M.; Yeguas-\linebreak Bolívar, E.; Calabrò, G.; Zingoni, A.; Taborri, J.}
        }
}

\address{%
$^{1}$ \quad {Department} 
 of Economics, Engineering, Society and Business Organization, University of Tuscia, {01100}~Viterbo,  Italy;  gianluca.morciano@unitus.it (G.M.); jose.alcalde@unitus.it (J.M.A.-L.); giuseppe.calabro@unitus.it (G.C.); andrea.zingoni@unitus.it (A.Z.)\\
$^{2}$ \quad Computing and Numerical Analysis, University of Córdoba, {14071} Córdoba, Spain; eyeguas@uco.es}

\corres{Correspondence:  michele.materazzini@unitus.it {(M.M.)}; juri.taborri@unitus.it (J.T.);\linebreak   Tel.: +39-328-0849306 (M.M.)
}

\abstract{
This study explores the use of virtual reality (VR) and artificial intelligence (AI) to predict the presence of dyslexia in Italian and Spanish university students. In particular, the research investigates whether VR-derived data from {Silent Reading (SR) tests}   and self-esteem assessments can differentiate between students that are affected by dyslexia and students that are not, employing machine learning (ML) algorithms. Participants completed VR-based tasks measuring reading performance and self-esteem. A preliminary statistical analysis (\emph{t}-tests and Mann--Whitney tests) on these data was performed, to compare the obtained scores between individuals with and without dyslexia, revealing significant differences in completion time for the SR test, but not in accuracy, nor in self-esteem. Then, supervised ML models were trained and tested, demonstrating an ability to classify the presence/absence of dyslexia with an accuracy of 87.5\% for Italian, 66.6\% for Spanish, and 75.0\% for the pooled group. These findings suggest that VR and ML can effectively be used as supporting tools for assessing dyslexia, particularly by capturing differences in task completion speed, but language-specific factors may influence classification accuracy.
}


\begin{document}


\section{Introduction}

The World Health Organization classifies specific learning disorders (SLDs) as neuro-developmental conditions characterized by persistent and significant difficulties in core learning abilities, such as reading, writing, and mathematics~\cite{who}. Students with SLDs, including dyslexia, dyscalculia, dysorthography, and dysgraphia, often face academic challenges, which can undermine their confidence levels~\cite{Ilaria2022}. Focusing on dyslexia, it varies by language due to differences in spelling rules. In transparent languages (e.g., Spanish, Italian), reading is slower but with fewer errors. In opaque languages (e.g., English, French), irregular spelling makes word recognition and spelling harder. Morphology also plays a key role in some languages, affecting how dyslexia is diagnosed and \mbox{supported~\cite{moore2023cross,hengeveld2018transparent,cossu1999acquisition}}.
In addition, beyond the primary difficulties associated with SLDs, secondary issues frequently arise, with low self-esteem being one of the most prominent. Self-esteem is a critical component of psychological well-being, influencing academic success, social \mbox{relationships~\cite{li2018social}}, and overall quality of life~\cite{kermode2001study}. Psychologists typically assess self-esteem using specialized questionnaires designed for this purpose. One of the most renowned and widely employed tools for such analysis is the Rosenberg Self-Esteem Scale (RSES)~\cite{donnellan2015measures}. This scale consists of 10 questions, with responses ranging from “strongly agree” to “strongly disagree”. The resulting scores offer valuable insights into individuals’ self-perception across various life domains~\cite{rosenberg1965rosenberg}. The RSES is a widely used and validated tool, even if the presence of negative wording has been criticized, since it seemed to be linked to lower reliability in some studies~\cite{tinakon2012comparison}.

The relationship between SLDs and self-esteem is complex and widely explored in educational psychology. Research consistently indicates that students with SLDs, such as dyslexia and dyscalculia, tend to have lower self-esteem compared to their peers without these challenges~\cite{alesi2012self}. Individuals with SLDs may experience significant impacts on both academic and social self-esteem, with affected students often experiencing diminished confidence and a reduced sense of self-worth~\cite{conley2007general}. Social dynamics further exacerbate these challenges, as peer rejection commonly leads to loneliness and feelings of inadequacy~\cite{parshurami2015study}. Additionally, anxiety and depression frequently accompany SLDs, compounding self-esteem issues and deepening the emotional struggles of affected students~\cite{shah2019relationship}. Dyslexia can also negatively impact self-concept and can lead to learned helplessness in the face of academic \mbox{difficulties~\cite{burden2005factors}}. Despite this general trend, not all studies report uniformly low self-esteem among students with SLDs. Some students with learning disabilities maintain positive self-esteem, pointing to the influence of factors such as personal resilience and social support. This variability highlights the multifaceted nature of self-esteem in students with SLDs and suggests that external and individual factors can significantly mitigate the negative psychological impact of learning challenges~\cite{pathrikar2016study}. As a result of the above, early identification of dyslexia is crucial for effective intervention; in fact, research shows that children identified as at risk for dyslexia can achieve significant progress when they are detected early and enrolled in targeted intervention programs~\cite{catts2018early}. 

 Traditional dyslexia tests involve reading meaningful and meaningless words aloud, while newer methods focus on Silent Reading (SR), missing spaces, and recognizing misspelled words. Digital technologies have enhanced dyslexia diagnosis through neurological data analysis (magnetic resonance imaging (MRI), functional magnetic resonance imaging (fMRI), electroencephalogram (EEG), and eye-tracking) and AI-driven test optimization. AI helps refine test selection, analyze results, and develop automated predictors~\cite{zingoni2021investigating}, achieving over 90\% accuracy in some studies~\cite{vrailexiapredictor1,morciano2024use}. Among  traditional tests, it is interesting to explore the reading preferences of the population. Research indicates that dyslexic children show little difference in performance between SR and reading aloud~\cite{van2022same}. Dyslexic adolescents, however, often prefer reading aloud, to enhance text comprehension~\cite{smyrnakis2021silent}, whereas dyslexic adults demonstrate limited improvement in reading speed during SR~\cite{gagliano2015silent}. Additionally, students struggle with SR due to difficulty with complex words, vocabulary, and focus, and often lack opportunities to develop independent reading skills~\cite{hairrell2010independent}.
 
It is important to note that traditional tools, such as the Rosenberg Self-Esteem Scale (RSES) and SR test, may lack the ecological validity and sensitivity needed to fully capture the nuanced experiences of children with SLDs~\cite{grigorenko2020understanding}. In order to generalize these tests to real-life situations, Virtual Reality (VR) has emerged as a promising tool for improving the ecological validity of neuro-psychological tests by providing more realistic testing environments than traditional methods~\cite{kourtesis2021validation}. Through interactive and immersive scenarios in a VR environment, users can engage in tasks and experiences that mirror real-life situations. This approach provides a more authentic and nuanced assessment of their self-esteem and emotional well-being~\cite{servotte2020virtual}. In addition, VR has proven to be an extremely effective tool in this field, where it has already been successfully applied as an effective educational tool for supporting students with learning disabilities~\cite{lin2024impact,drigas2022virtual}, stimulating empathy~\cite{alcalde2023vr,alcalde2025} and self-esteem~\cite{pijnenborg2022discovr}. VR has also been successfully used on multiple fronts applied to dyslexic children, these include teaching~\cite{maskati2021using}, cognitive rehabilitation~\cite{maresca2022use,maresca2024effectiveness}, and even in diagnosis, when used in combination with machine learning and eye-tracking devices, achieving 98\% accuracy~\cite{vaitheeshwari2024dyslexia}. Furthermore, VR aims to increase engagement~\cite{lin2024impact}, accessibility~\cite{chalkiadakis2024impact}, and concentration~\cite{ko2020effects}, by providing an interactive experience that builds on the strengths of modern devices. Lastly, VR offers the advantage of enabling the rapid, repeatable, and efficient collection of large amounts of data during user interactions within the VR environment~\cite{wen2024vr}, and the immersive nature of VR allows for real-time tracking of user responses~\cite{spitzley2019feasibility}, reducing manual evaluation time, while ensuring a more structured and reliable dataset. Given these findings, this work explores the potential of combining self-esteem assessment with SR evaluations as an additional diagnostic tool for dyslexia, all within an immersive VR environment.

Collected data can be analyzed using Artificial Intelligence (AI), particularly through Machine Learning (ML), creating a synergistic integration of these two cutting-edge technologies. Several studies have already demonstrated the efficiency of various ML algorithms for the detection of learning disorders. For example with k-Nearest Neighbors (k-NNs)~\cite{khan2018machine}, where a machine learning system was developed to identify dyslexia risk based on data from 857 first-grade students in Malaysia. The dataset, collected in a prior study using specially designed tests, was manually labeled by a dyslexia expert and preprocessed to ensure quality, and trained using a 70/30 train--test split. The model achieved 99\% accuracy in classification, with validation against expert assessments confirming its reliability. 
Another study adopted a Support Vector Machine (SVM)~\cite{tamboer2016machine}, where an SVM-based anatomical classifier was used to distinguish students with (22) and without (27) dyslexia, achieving 80\% accuracy in the training sample, but dropping to 59\% in a general population sample of 876 subjects. Key brain regions involved were the Left Inferior Parietal Lobule (LIPL) and the Left and Right Orbitofrontal Gyrus (LOFG, ROFG). Brain structure correlated with dyslexia severity. However, the high false positives in the second sample suggest the classifier may capture broader anatomical variations rather than dyslexia-specific traits. 
A further work applied Logistic Regression (LR)~\cite{plonski2014dealing}, the study analyzed neuro-imaging data from 130 dyslexic and 106 control subjects, each described by 742 cortical features. A LR classifier initially achieved an Area Under Curve  (AUC) of 0.61, but step-forward feature selection improved the performance to 0.73 using only 25 features (3.4\% of the total). To address the site dependency, two methods were proposed: site-dependent whitening (SDW) and site-dependent extension (SDE). Both methods enhanced classification performance to AUC 0.82 (SDW) and 0.83 (SDE), representing a 12\% improvement over naive feature selection and a 25\% improvement over using all features. 
Finally, a combination of SVM, LR, and Random Forest (RF) can be find in the literature~\cite{plonski2017multi}; the study examined grey matter differences between typical and dyslexic children using a multivariate approach on T1-weighted MRI images from 236 children across Poland, France, and Germany. Participants aged 8.5--13.7 were screened for dyslexia based on standardized reading tests, IQ, and exclusion of ADHD or neurological disorders. The classification of dyslexic vs. control subjects achieved moderate accuracy (AUC = 0.66, accuracy (ACC) = 0.65 with 10-fold cross-validation) using LR, SVM, and RF classifiers. Dyslexic children showed higher mean curvature and greater folding index in left-hemisphere language-related regions, supporting prior findings of cortical folding anomalies in dyslexia. While multi-site variability and social background differences pose challenges, these results suggest geometric cortical properties as potential biomarkers of dyslexia.

From this perspective, it is clear how the combination of VR scenarios to gather data and ML algorithms for data processing could represent a groundbreaking approach, enhancing the use of technology to complement traditional methods for diagnosing dyslexia. To the best of the authors' knowledge, while various efforts have been made to incorporate VR and ML tools into the clinical assessment of learning disorders, the virtualization of SR and self-esteem tests remains unexplored. Similarly, the use of machine-learning algorithms to automatically identify the presence of dyslexia based on data gathered from the abovementioned tests is still an open challenge. {Recent metric-learning research shows that block-level feature augmentation combined with self-supervised auxiliary loss can markedly improve few-shot classification under extreme data scarcity, a result that directly motivated the present exploratory design~\cite{zhang2024sample}.}

This study introduces an innovative VR-based tool designed to administer both SR and Rosenberg self-esteem tests. The tool collects digital data that can be used to train ML algorithms to identify the presence of dyslexia. As an additional innovative aspect, this study adopted a cross-linguistic approach by exploring how the novel methodology could be applied to data collected from both Italian and Spanish students.

\section{Materials and Methods}

\subsection{Setup and Virtual Test}

The study  was conducted using Meta Quest 2 {(Meta Platforms, Inc.: Menlo Park, CA, USA)} head-mounted displays (HMDs) with the “Out of the Box” app pre-installed. This application, developed through collaboration between experts in educational psychology and VR technologies, was designed in~\cite{yeguas2022determining} and developed in~\cite{materazzini24} as part of the VRAIlexia project~\cite{vrailexiasummary}. The virtual environment was designed to replicate real-world challenges faced by students with specific learning disorders (SLDs) and included two primary assessments. 

The first was a SR evaluation, measuring reading-related cognitive and motor performance through interactive tasks. This is a commonly used test~\cite{santulli2018superreading} where participants are asked to read a text in which are described various task to be accomplished. The VR SR scene reproduces the reading-comprehension task of the BDA 16--30
 adult dyslexia battery {(Giunti Psychometrics: Florence, Italy)}~\cite{bda}, substituting the physical three-color button panel and sheets with  virtual ones, while leaving the stimulus--response format unchanged. During the reading, users were asked to select buttons based on color, follow specific button sequences, hold and release buttons as instructed, choose words within a text, and use voice recognition to complete verbal interactions. The system recorded various data points, including the start time, error count, interaction duration, environmental factors, and voice recognition errors. 

The second assessment was the Rosenberg Self-Esteem Test, a well-established \mbox{10-item} scale evaluating global self-esteem in areas such as academic performance, social relationships, and personal competence. Participants responded to each statement using a four-point Likert scale ranging from “strongly agree” to “strongly disagree.” Data collection included start time, test duration, environmental context, and emotional responses. Self-esteem levels were categorized based on predefined cut-off points, with scores ranging from 30 to 40 indicating high self-esteem, 26 to 29 representing a medium level, and scores below 25 classified as low self-esteem. Responses were further analyzed in relation to emotional state descriptors relevant to students with SLDs, providing a deeper understanding of their psychological well-being. The ability to collect detailed behavioral insights within a controlled VR setting allows for more precise assessment and the development of targeted intervention strategies tailored to individual needs.

\subsection{Participants and Experimental Protocols}

A total of 80 university students participated in this study, divided into two independent samples of 40 students each. The first sample consisted of 40 Italian university students, half of whom were diagnosed with an SLD (SLD group), of whom 100\% had dyslexia and 85\% also had mixed-type, whereas the remaining 20 formed the control group (CG) with no diagnosed disorders. These participants were recruited from the University of Tuscia and the University of Perugia. The second sample included 40 Spanish university students, similarly divided into 20 with SLDs (of whom 100\% had dyslexia and 55\% also had mixed-type) and 20 in the control group. These participants were recruited from the University of Córdoba under the same selection criteria. {All dyslexic participants held a current clinical certificate issued by a licensed neuro-psychologist; these certificates served as the ground-truth labels for supervised machine learning.} For both groups, the inclusion criteria required participants to be university students over 18 years of age, native speakers of either Italian or Spanish, and to have no physical conditions that could interfere with VR use. Additionally, participants were excluded if they had any cognitive or psychological disorders beyond SLDs, ensuring a more homogeneous sample for comparison.
{Participation was entirely voluntary and no monetary or material compensation was offered, and behavioral data were collected in fully anonymous form.}

{In Table~\ref{tab:demo}, we provide descriptive information for each subgroup, including mean age, gender distribution, and academic background.}

\begin{table}[H]
	\centering
	\caption{Demographic characteristics of the participant subgroups.}
	
	\label{tab:demo}
	\begin{tabularx}{\textwidth}{LCCc}
		\toprule
		\textbf{Group} & \textbf{Mean Age}  & \textbf{Gender (M/F)} & \textbf{Field of Study} \\ 
		\midrule
		Italian CG & 25.8 $\pm$ 2.8 & 11/9 & Engineering--Humanities \\
		Italian SLD  & 21.6 $\pm$  2.8 & 10/10 & Engineering--Humanities \\
		Spanish CG & 26.0 $\pm$  3.1 & 12/8 & Computer Science--Nursing\\
		Spanish SLD  & 24.9 $\pm$  4.7 & 10/10 & Computer Science--Nursing \\ 
		\bottomrule
	\end{tabularx}
\end{table}

\vspace{-3pt}

\begin{figure}[H]
	\subfloat[][\centering]{\includegraphics[width=0.3\textwidth]{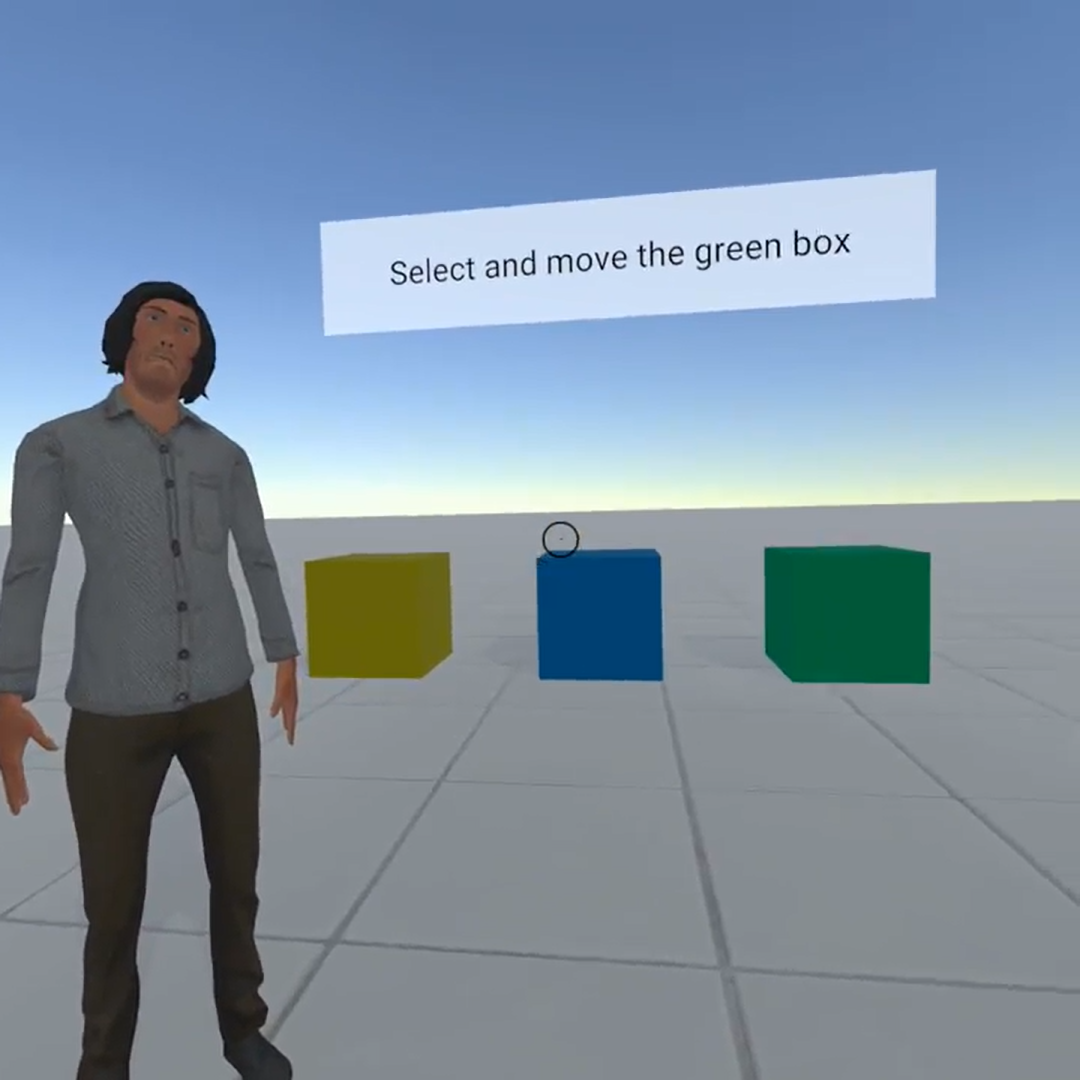}\label{subfig:Select}} \quad
	\subfloat[][\centering]{\includegraphics[width=0.3\textwidth]{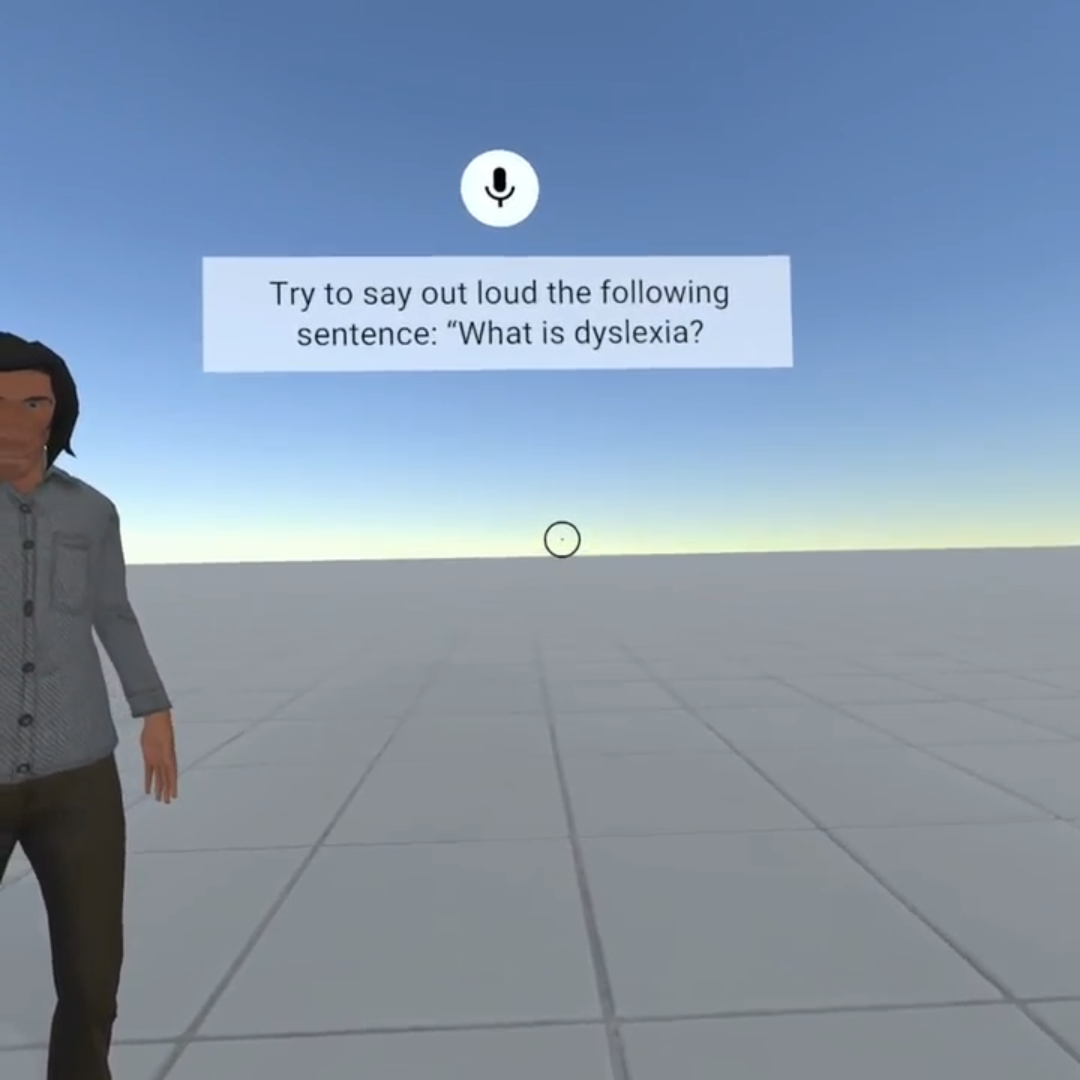}\label{subfig:Speech}} \quad
    \subfloat[][\centering]{\includegraphics[width=0.3\textwidth]{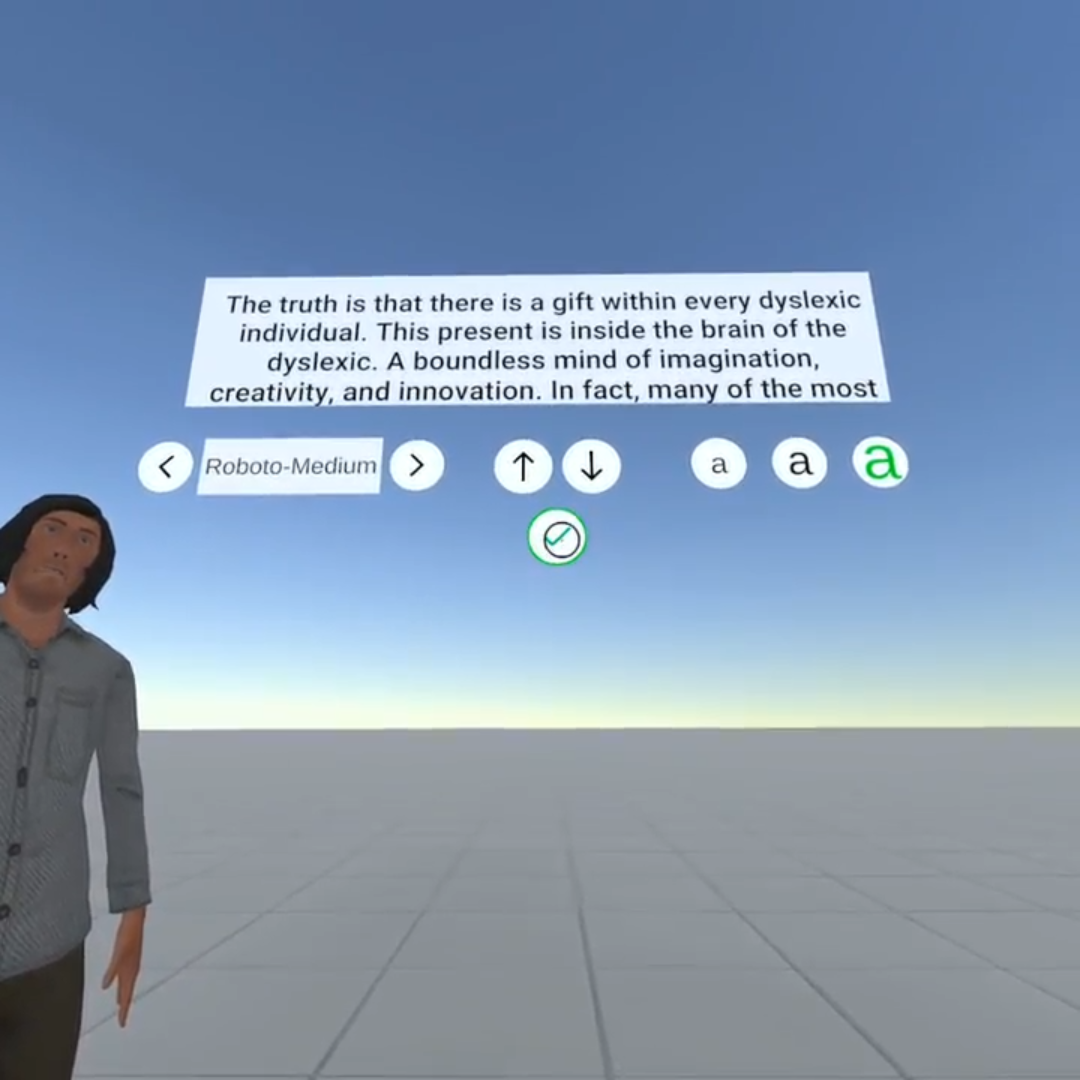}\label{subfig:Customize}} \\
    \caption{{Screenshots} 
 of the familiarization phase. {(\textbf{a}) How to interact; (\textbf{b}) Test of the speech-to-text algorithm; (\textbf{c}) Font customization.} 
}
    \label{fig:Intro1}
\end{figure}

Each student was instructed to wear the HMD and complete both tests. Upon launching the application, users were greeted by an initial screen where they provided basic socio-demographic information, including age, gender, and whether they had been diagnosed with one or more specific SLDs. To help participants familiarize with the VR environment, a virtual character introduced the experience and guided them through the interaction mechanics. Given the study’s research objectives, user movement within the virtual space was deliberately restricted, to ensure standardized conditions across all participants. Instead of freely navigating the environment or using hand gestures for selection, users interacted via head movement. A small circle at the center of the screen served as a pointer, requiring participants to align it with an interactive element before confirming their choice using the controller buttons, {Figure} ~\ref{fig:Intro1}a. This constraint ensured uniform interaction patterns, reducing the variability in data collection, while maintaining focus on the cognitive processes involved in reading and response selection.
For the SR scenario, voice interaction was introduced as an additional mode of engagement. The virtual character explained this feature by instructing participants to repeat a specific sentence, familiarizing them with the speech-to-text algorithm Figure~\ref{fig:Intro1}b. An audible signal confirmed when the system had successfully processed the spoken input. Since the study targeted individuals with reading disorders, the application also allowed users to adjust text size and font to accommodate their reading preferences and minimize accessibility barriers, Figure~\ref{fig:Intro1}c. {This familiarization ensured that the VR interface, unfamiliar yet identical for all participants, did not bias the subsequent SR measures.}

\begin{figure}[H]
	\subfloat[][\centering]{\includegraphics[width=0.3\textwidth]{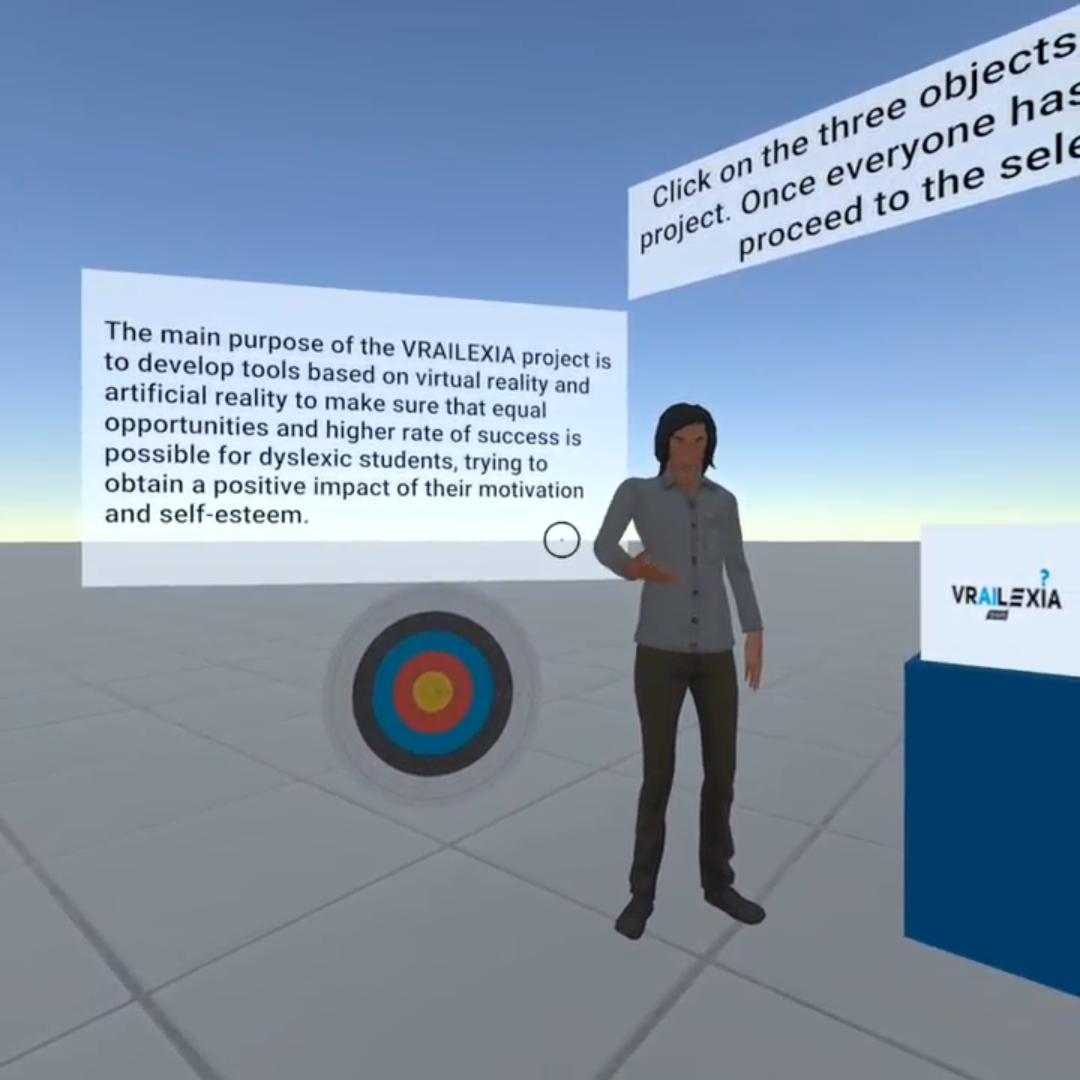}\label{subfig:Expl2}} \quad
	\subfloat[][\centering]{\includegraphics[width=0.3\textwidth]{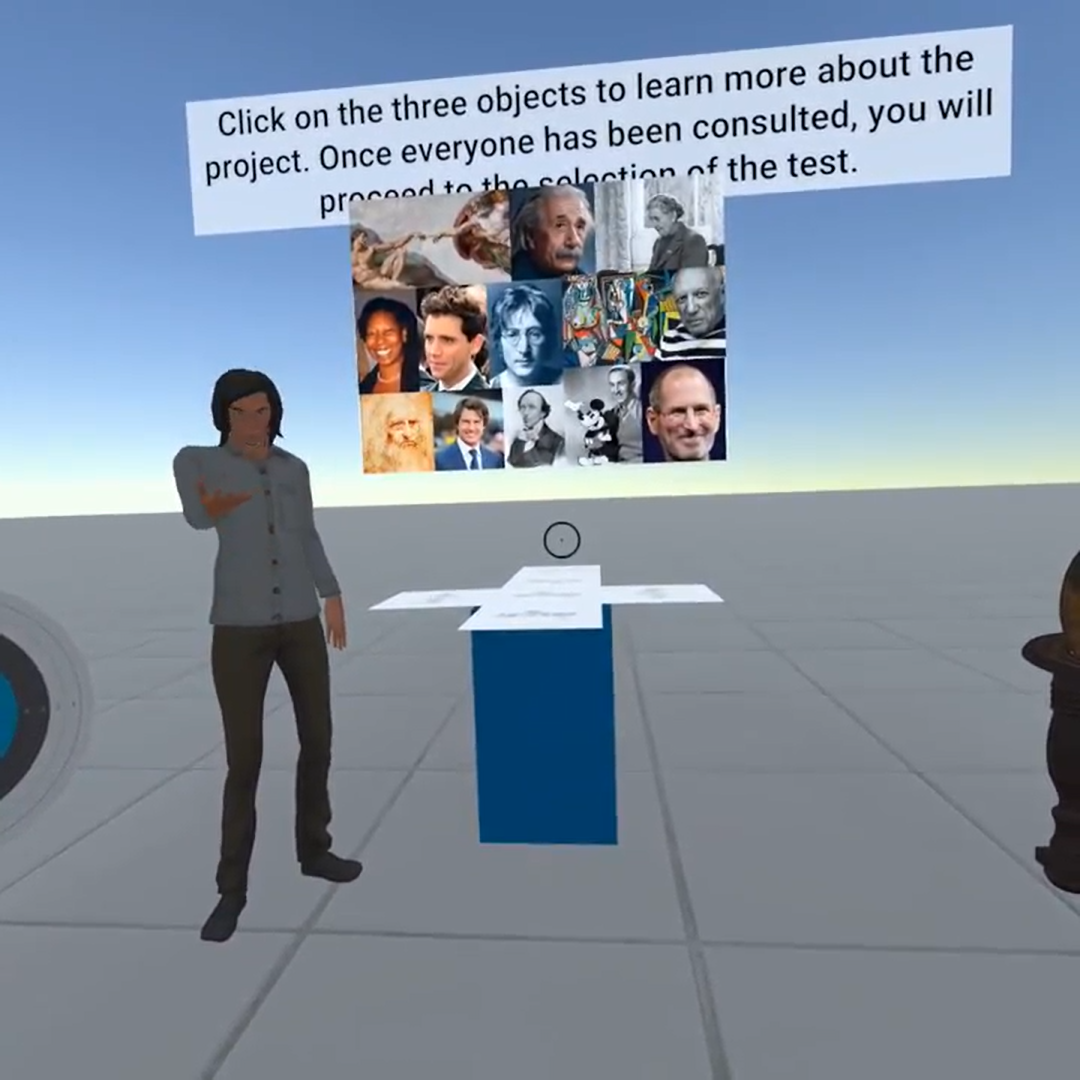}\label{subfig:Expl1}} \quad
    \subfloat[][\centering]{\includegraphics[width=0.3\textwidth]{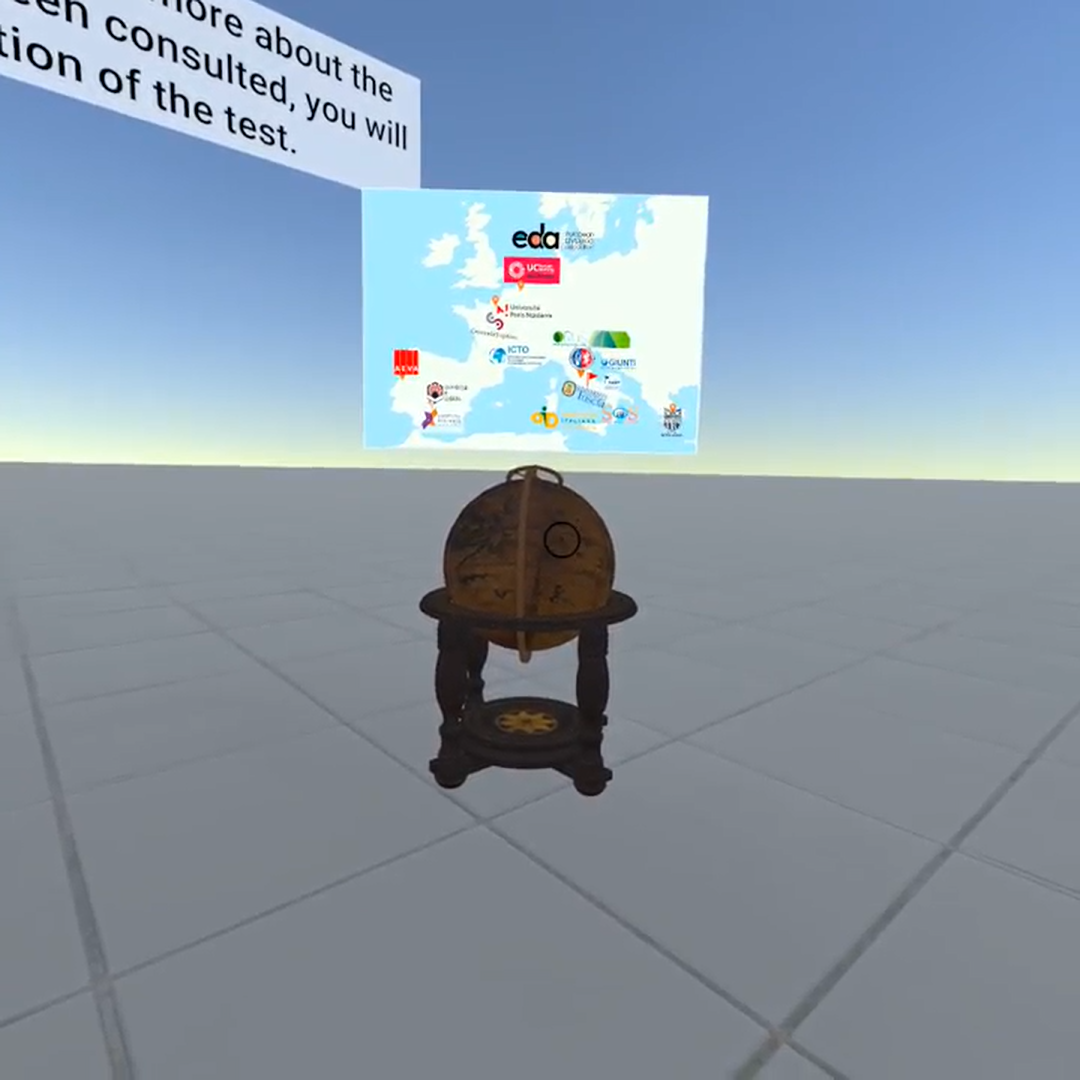}\label{subfig:Expl3}} \\
    \caption{{Screenshots} of the educational segment.  {(\textbf{a}) VRAIlexia's project explanation; (\textbf{b}) well-known figures with certified SLDs; (\textbf{c}) overview of institutions participating in the project.}}
    \label{fig:Intro2}
\end{figure}

Once participants had become comfortable with the VR interaction mechanics, the virtual character provided an educational segment on SLDs and the VRAIlexia project~\cite{vrailexia} \mbox{Figure~\ref{fig:Intro2}a}. This segment aimed to increase awareness of reading difficulties and foster a sense of relatability by presenting a list of well-known historical and contemporary figures diagnosed with learning disorders, Figure~\ref{fig:Intro2}b. Lastly, all the actors that participated in the development of the VRAIlexia project were shown, Figure~\ref{fig:Intro2}c. By incorporating this step, the experience was designed to be more engaging, reducing the potential anxiety associated with performance-based assessments.\vspace{-6pt}

\begin{figure}[H]
	\subfloat[][\centering]{\includegraphics[width=0.3\textwidth]{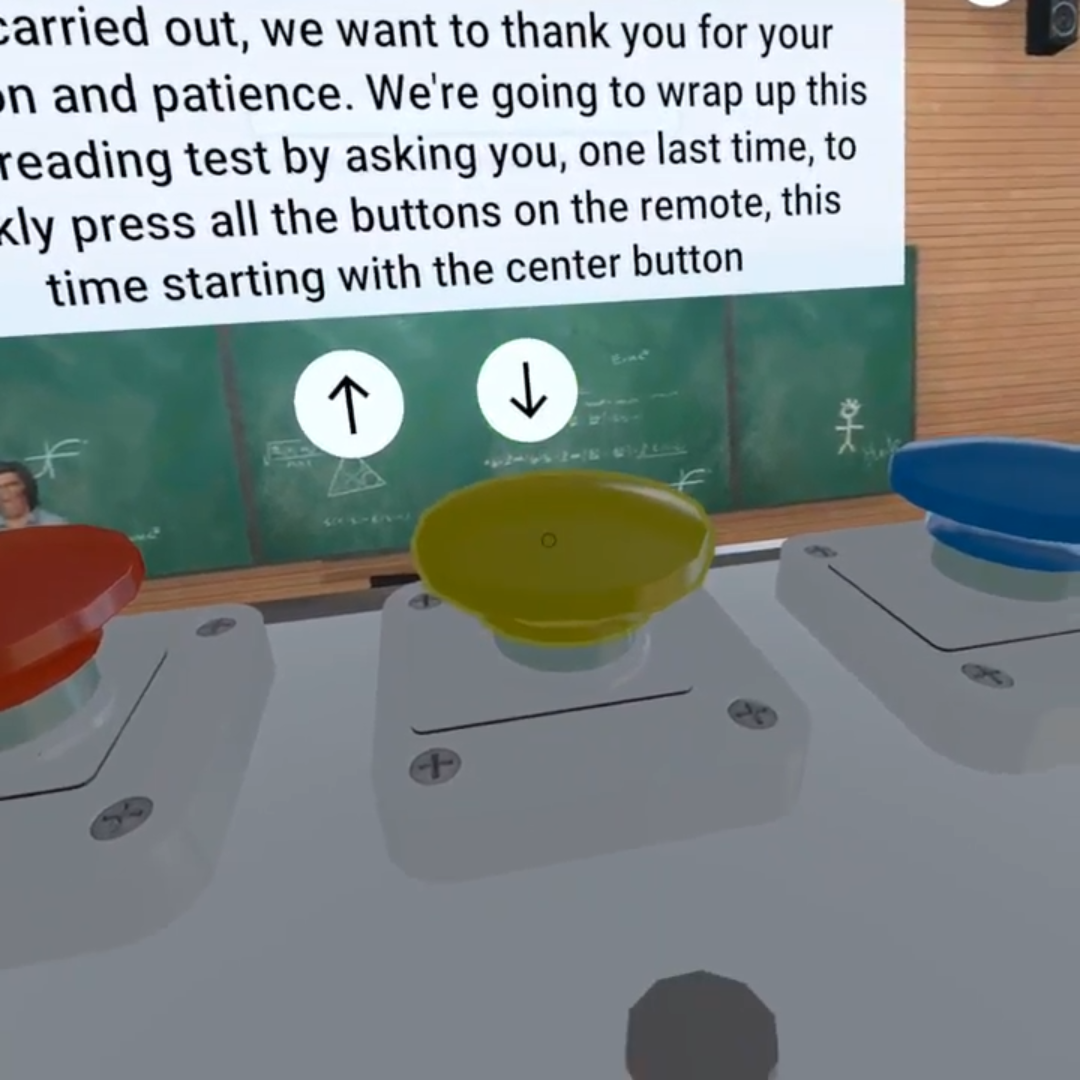}\label{subfig:SR}} \quad
	\subfloat[][\centering]{\includegraphics[width=0.3\textwidth]{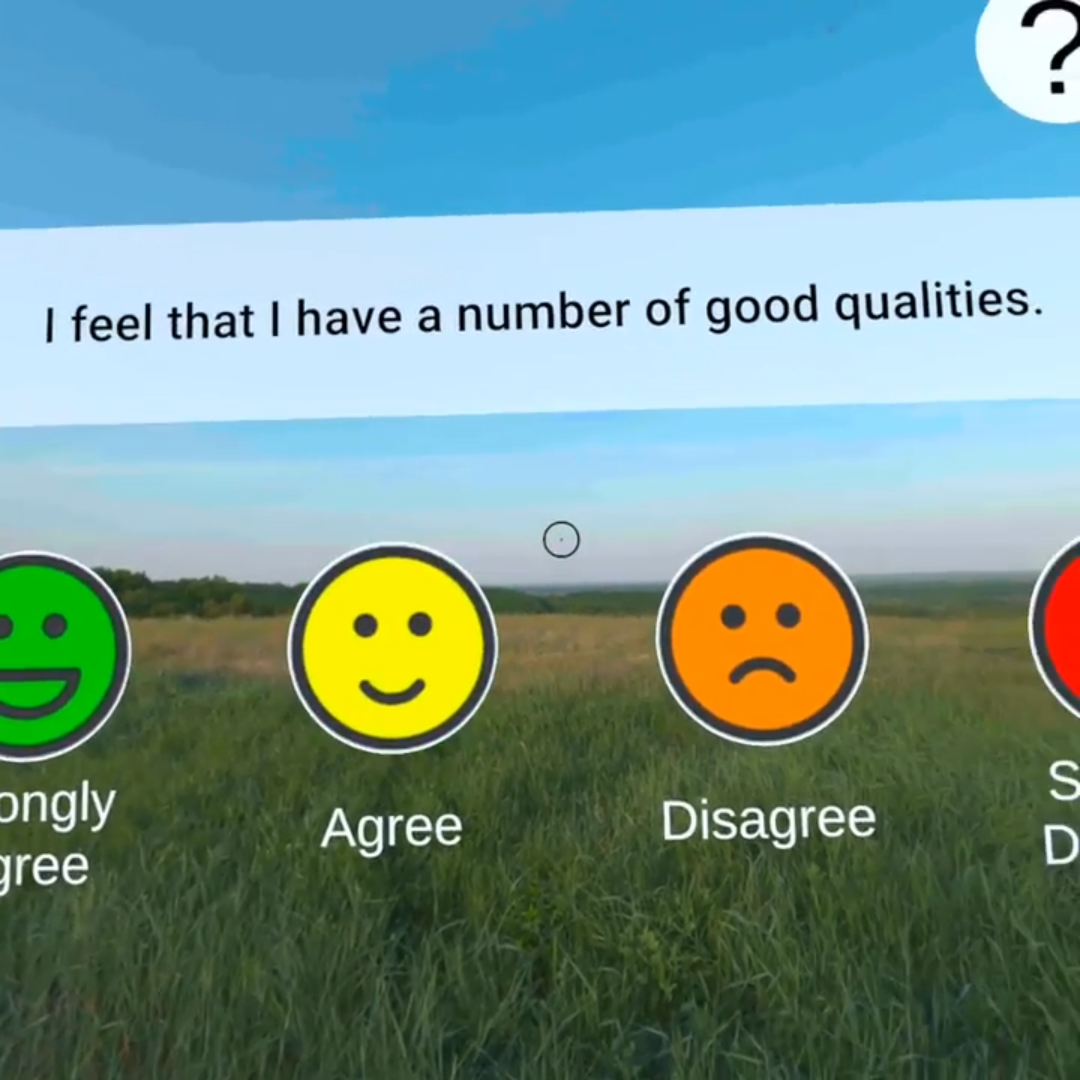}\label{subfig:RSES}} \quad
    \subfloat[][\centering]{\includegraphics[width=0.3\textwidth]{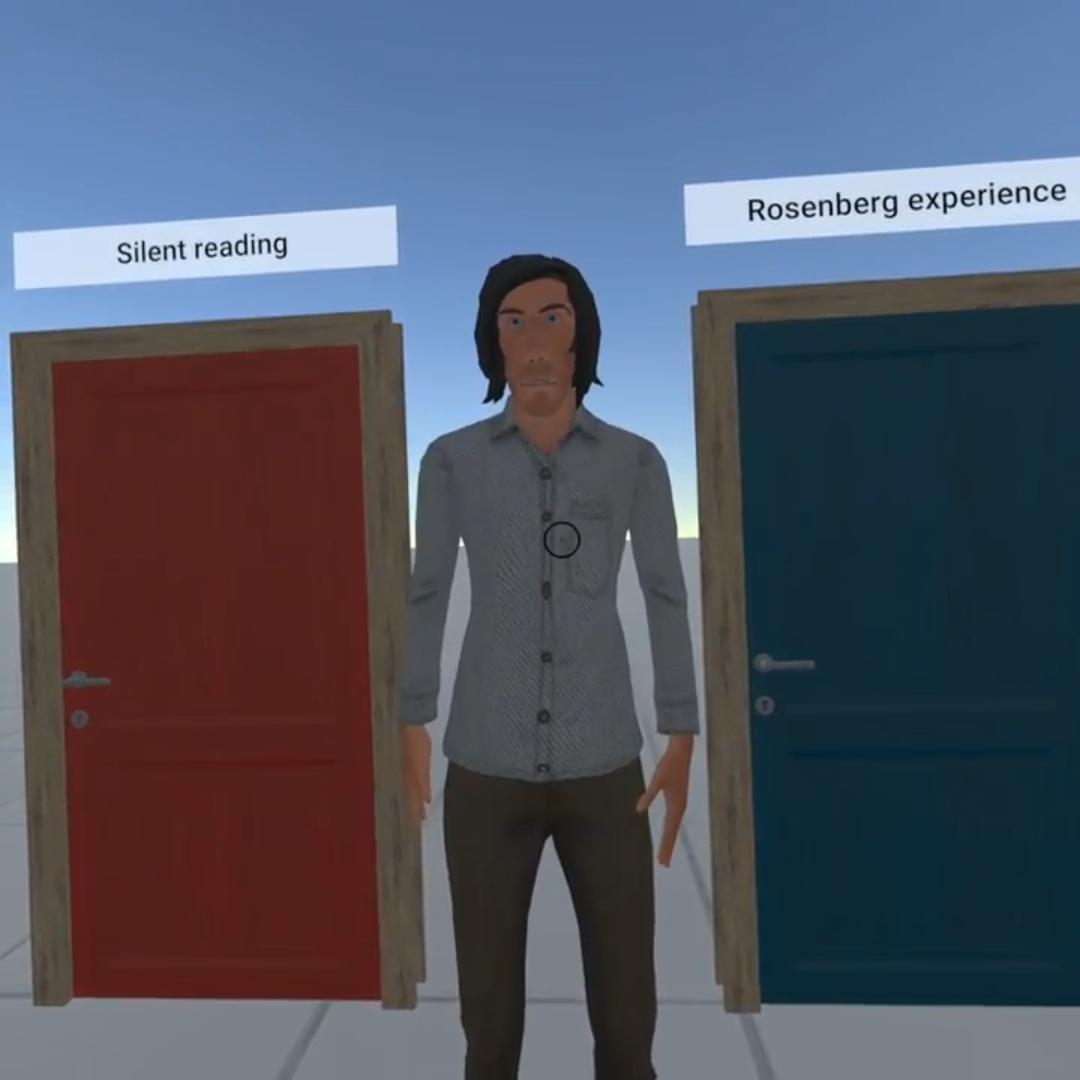}\label{subfig:Doors}} \\
    \caption{{Screenshots}  of the tests phase.  {(\textbf{a}) SR; (\textbf{b}) RSES; (\textbf{c}) Opportunity to choose which test to start with.} 
 }
    \label{fig:Intro3}
\end{figure}

The SR test was structured to evaluate participants’ ability to process written content while simultaneously engaging in interactive tasks. Because selections were made through head movement, users often had to shift their focus between the text and response options, simulating real-world reading challenges experienced by individuals with SLDs, Figure~\ref{fig:Intro3}a. This additional cognitive load was an intentional design choice, reflecting the difficulty many dyslexic readers face when needing to process visual information while performing secondary tasks. Participants were required to interact five times with three colored buttons at the bottom of the screen, select a specific word within the text and verbally repeat three words from the passage. These tasks were designed to assess different aspects of reading engagement, such as visual tracking, word recognition, and short-term verbal recall. Following the reading assessment, the Rosenberg Self-Esteem Test was administered using the same interaction mechanics. Participants responded to ten statements assessing global self-esteem, selecting their answers by aligning the central pointer with the desired option, before confirming with the controller Figure~\ref{fig:Intro3}b. Unlike a fixed testing order, users were given the freedom to choose which test to complete first, allowing them to follow their preferred sequence, Figure~\ref{fig:Intro3}c. This approach aimed to reduce stress and increase participant comfort, ensuring that individual preferences were accounted for in the testing process. Additionally, it allowed avoiding bias in the results due to the test sequence.

{All the operations described above are part of an experimental protocol that has been approved by the Ethical Committee  CEIm Provincial de Córdoba n. 367 (27/11/2024).}\vspace{-8pt}

\subsection{Data Analysis}

To assess the effectiveness of the VR-based psychometric tests, we implemented a supervised {ML classification framework}. The dataset consisted of performance metrics extracted from VR assessments, including error rates, total response times, and \mbox{self-esteem scores}. 

\begin{figure}[H]
	\includegraphics[width= 1\textwidth]{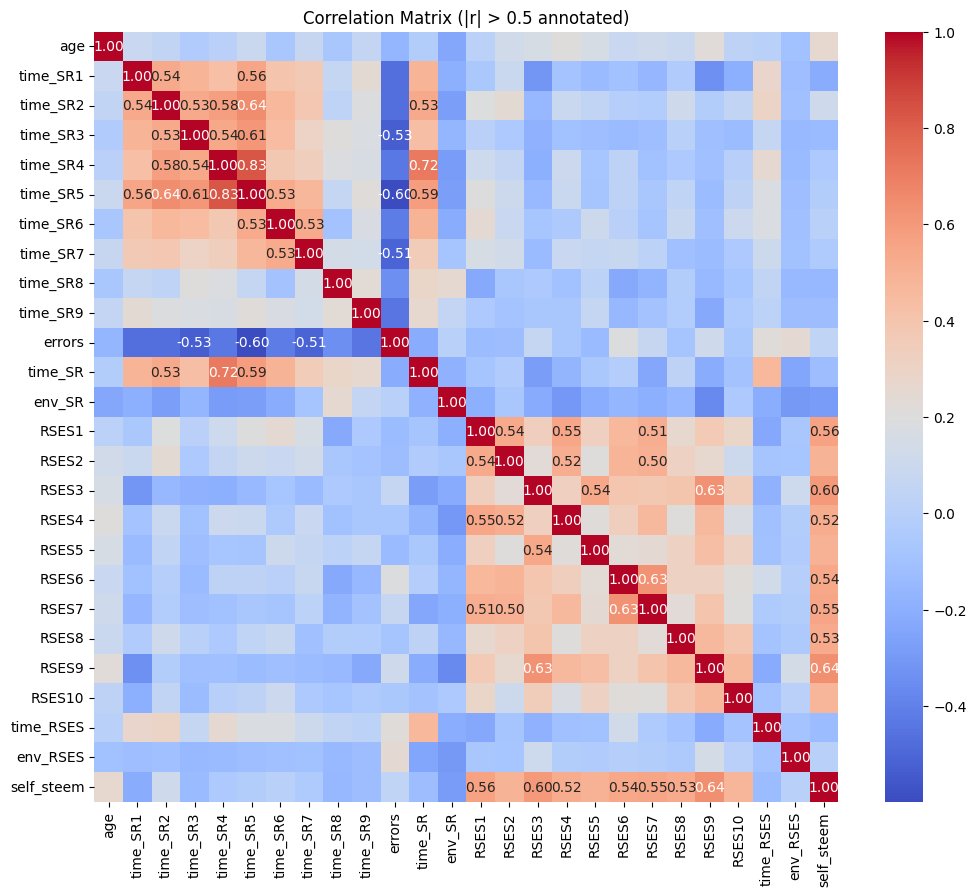}
	\caption{{Correlation} matrix.}
	\label{fig:CM}
\end{figure}

{To reduce dimensionality and remove redundant information, a correlation matrix was computed among the numerical initial features. As shown in Figure~\ref{fig:CM}, we observed high correlations among the individual time measurements of SR (e.g., time\_SR1 to time\_SR9) and RSES (e.g., RSES1 to RSES10) items, as well as between these and their respective totals. Therefore, we retained only the total SR time and total RSES time as synthetic indicators. Similarly, individual RSES responses showed high internal correlations and no particular item demonstrated discriminative power over the others. As a result, only the global self-esteem score was kept. The same rationale was applied to the error-related variables, where only the total number of SR errors was selected as a compact and meaningful metric. This resulted in a final feature set composed of four variables: SR errors, total SR time, total RSES time, and self-esteem score.}

The target variable was the presence of dyslexia, which was binarized into class labels (0: no dyslexia, 1: dyslexia).
Prior to model training, the dataset was preprocessed to remove missing values. The dataset consisted of 41  features {(s}ee Table~\ref{tab:feature_summary}) extracted using the Out of the Box application, which synthesized the results from two psychometric tests. These features included the number of errors in the different stages of the SR, as well as the Rosenberg test score for each of the items. Additionally, a correlation study was conducted to identify the most relevant features for dyslexia prediction and to avoid redundant information among features. Based on this analysis, the final feature set was reduced to four key predictive variables: errors, total SR time, total RSES time, and self-esteem score. Finally, the dataset was then split into training and testing sets using an {80--20}  ratio, using stratified sampling to maintain the class distribution.

 A set of widely used supervised learning algorithms were evaluated for dyslexia prediction. To optimize their predictive performance, a systematic search over predefined hyperparameter grids was conducted. The models evaluated and their corresponding hyperparameter tuning strategies were as follows:

\begin{itemize}
    \item	{LR}: The hyperparameter optimization process evaluated two types of regularization: L1, which promotes sparse solutions by driving some coefficients to zero, and L2, which distributes regularization more evenly across all parameters. These were assessed in combination with two optimization algorithms: Limited-memory BFGS, a quasi-Newton method suited for smooth, differentiable objectives and typically used with L2 regularization; and Liblinear, a coordinate descent-based solver that efficiently handles both L1 and L2 penalties;
    \item	{SVM}: This function defining how data are mapped into a higher-dimensional space was explored using three types of kernels: a simple linear kernel, a polynomial kernel (\mbox{2nd degree}), and a RBF kernel, measuring similarity based on the distance between points and using a Gaussian function to assign higher weights to closer data points. The parameter controlling how the influence of individual points decreased with distance was evaluated with two predefined settings, one that scaled the values based on the reciprocal of the number of features in the dataset, and another that automatically adjusted based on the range of input values;
    \item	{k-NN}: The number of neighboring data points considered for classification was tested with values of 3, 5, and 7;
    \item	{DT}: The maximum depth allowed in the tree was varied between 10 and 15. The method for determining the best split at each node was tested using two well-known criteria: gini and entropy;
    \item	{RF}: The number of decision trees combined in the ensemble was tested with values 10, 20, 30, and 40. The other parameters considered were the same as those for the Decision Tree Classifier.\vspace{-8pt}
\end{itemize}

\begin{table}[H]
	\caption{Features extracted from the Out of the Box application.}
	
\begin{adjustwidth}{-\extralength}{0cm}
\begin{minipage}{\fulllength}
\label{tab:feature_summary}

	\begin{tabularx}{\textwidth}{
		>{\hsize=0.8\hsize}X 
		>{\hsize=0.5\hsize}X 
		>{\hsize=0.5\hsize}X 
		>{\hsize=2.2\hsize}X}
		\toprule
		\textbf{Feature Name} & \textbf{Unit} & \textbf{Test Origin} & \textbf{Description} \\
		\midrule
		Demographic info. & -- & General & Basic demographic information (age, sex, and language) \\
		Reported SLDs & -- & General & SLDs diagnosed to the participant (dyslexia, dyscalculia, dysgraphia, and dysorthography) \\
		Additional problems & -- & General & Other cognitive or developmental conditions (e.g., ADHD) \\
		Device & -- & General & Device used during the VR session (headset or cardboard) \\
		SR response times & Seconds & SR & Time taken to perform each of the nine comprehension tasks during the SR test \\
		SR accuracy & Boolean & SR & Whether each response in the reading task was correct \\
		Total reading time & Seconds & SR & Time to complete the entire reading assessment \\
		Total reading errors & Count & SR & Total number of incorrect answers in the reading test \\
		Environment noise & Boolean & SR & Quality of the environment during the reading task \\
		Microphone issues & Boolean & SR & Presence of technical microphone problems \\
		Self-esteem responses & Ordinal (1–4) & RSES & Responses to 10 items in the RSES \\
		Self-esteem score & Score (0–30) & RSES & Sum of Rosenberg items indicating global self-esteem \\
		Total RSES time & Seconds & RSES & Time taken to complete the self-esteem test \\
		Environment  & Categorical & RSES & Conditions during the self-esteem assessment \\
		\bottomrule
	\end{tabularx}
\end{minipage}
\end{adjustwidth}
\end{table}

\vspace{-3pt}

{The choice of traditional ML algorithms was guided by the dataset size and the nature of the input features. With four final predictors and a sample size of 80 participants, deep learning models such as RNNs or MLPs were unsuitable due to their higher risk of overfitting, increased computational demands, and lack of interpretability. In contrast, classical ML models provide robust performance in low-data regimes and are easier to interpret in terms of feature contributions and decision boundaries, making them preferable for an application with potential clinical and educational implications.}

Finally, model performance was assessed using standard classification metrics: accuracy and F1-score. Each model’s effectiveness was evaluated on the test set, and the results were compared to determine the most suitable approach for VR-based dyslexia assessment. To investigate potential cross-cultural differences and validate the robustness of the predictive models, the analysis was conducted in three phases: first, using data exclusively from Italian students, then, from Spanish students, and finally, using the combined dataset.

\subsection{Statistical Analysis}

In addition to the ML approach, statistical analyses were conducted to evaluate differences between the CG and SLD  groups for every sample of data (CG$_{i}$ and SLD$_{i}$ for the Italians, CG$_{s}$ and SLD$_{s}$ for the Spanish, and CG$_{p}$ and SLD$_{p}$ for the pooled group). To determine whether the time taken to complete both tests differed significantly between SLDs and CG, an independent \emph{t}-test {and a post hoc power analysis} were applied. Meanwhile, the non-parametric Mann--Whitney (MW) test was used to analyze the Rosenberg self-esteem scores and the number of errors made during the SR task. A significance level of $0.05$ was set for all statistical tests. 

\section{Results}

In this section, the results concerning the assessments are presented, first subdivided by nationality and finally as an uniform group.

\subsection{Italian Group}
\subsubsection{Differences Between Groups}

\begin{figure}[H]
	\subfloat[][\centering]{\includegraphics[width=0.48\textwidth]{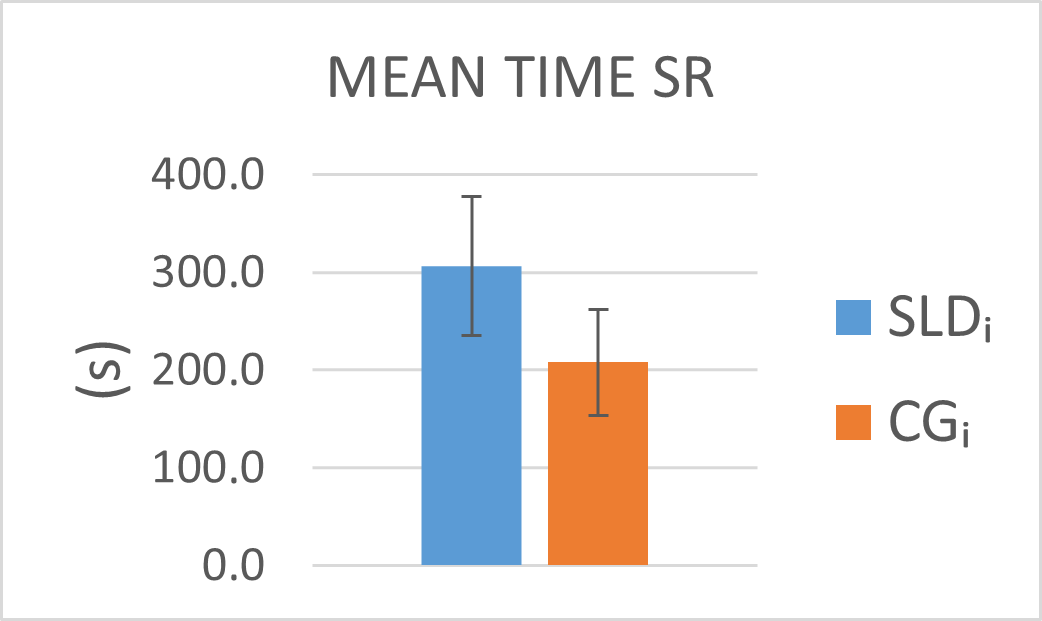}\label{subfig:MTSR_ITA}} \quad
	\subfloat[][\centering]{\includegraphics[width=0.48\textwidth]{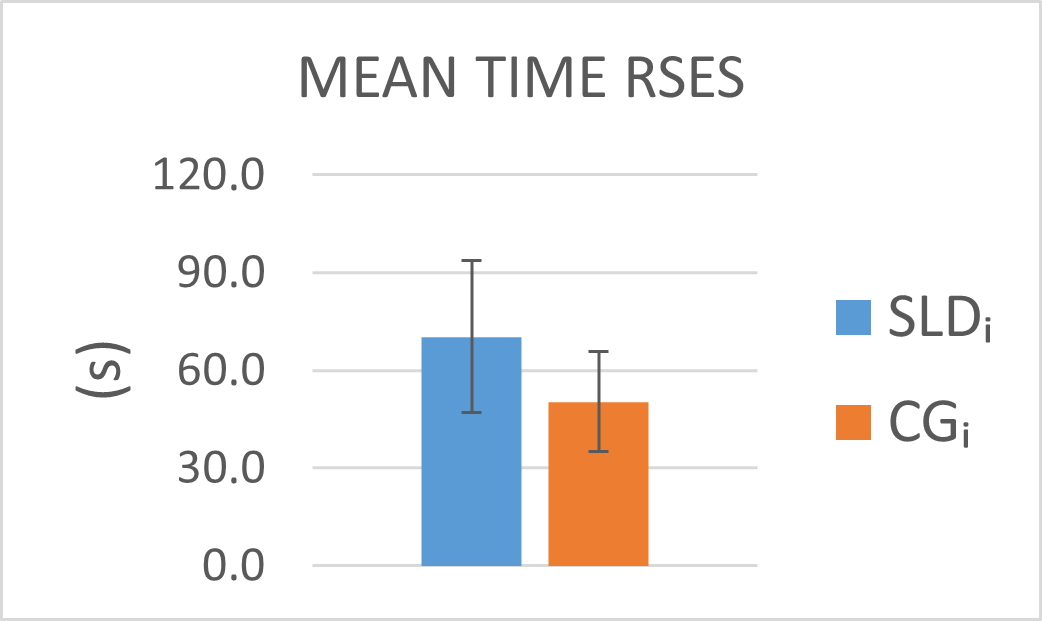}\label{subfig:MTRSES_ITA}}\\
    \subfloat[][\centering]{\includegraphics[width=0.48\textwidth]{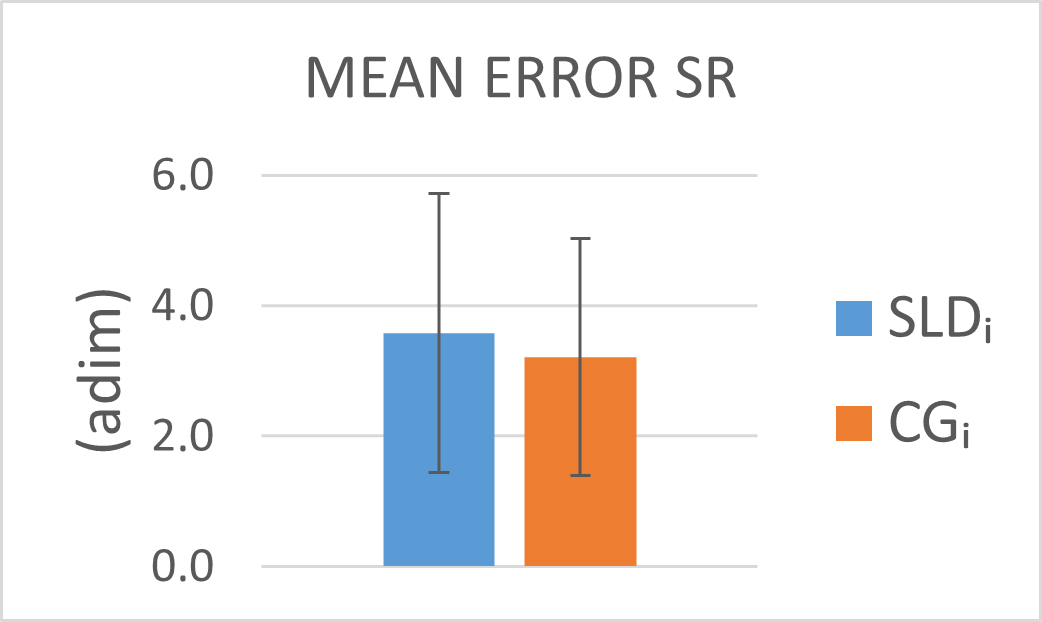}\label{subfig:MESR_ITA}} \quad
    \subfloat[][\centering]{\includegraphics[width=0.48\textwidth]{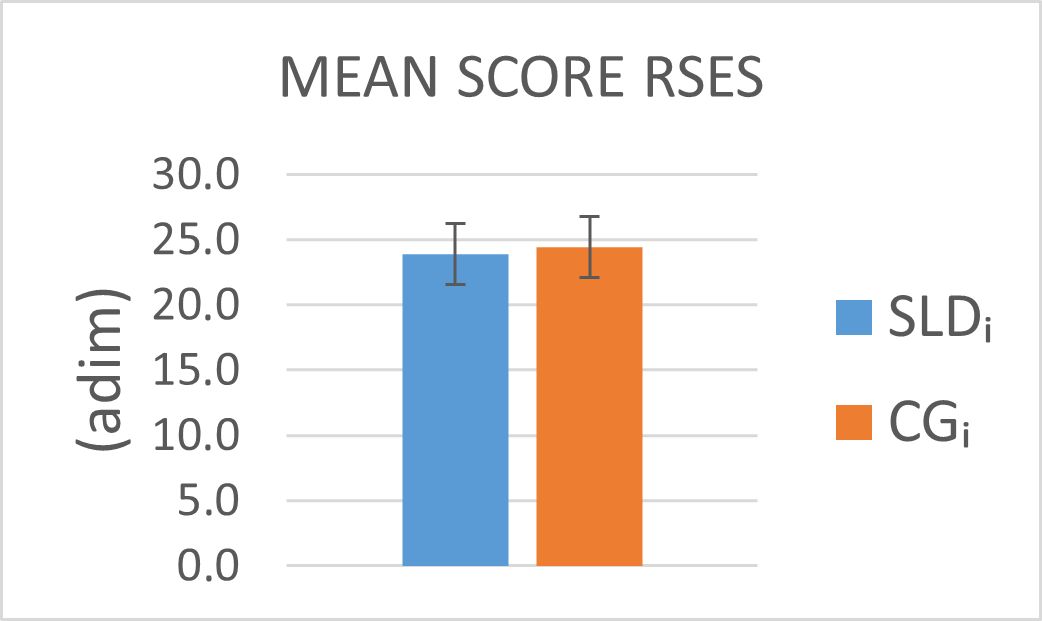}\label{subfig:MSRSES_ITA}}\\
    \caption{{Performance} of the Italian group: (\textbf{a}) average time to perform SR test; (\textbf{b}) average time to perform RSES test; (\textbf{c}) average errors made during SR test; (\textbf{d}) average of the scores obtained in the RSES test.}
    \label{fig:ItalianStat}
\end{figure}

As shown in Figure~\ref{fig:ItalianStat}, the results illustrate the differences in performance between the two Italian groups. From left to right, from top to bottom, the first set of data, \mbox{Figure \mbox{\ref{fig:ItalianStat}a}}, results indicate that the mean time to finish the SR test was 306 s for the SLD Italian group (SLD$_{i}$) and 200 s for the Italian CG (CG$_{i}$). The second set, Figure~\ref{fig:ItalianStat}b, represents the mean time required to complete the RSES test, with SLD$_{i}$ participants averaging 70.3 s and the CG$_{i}$ completing it in 50.3 s. The third graphic, Figure~\ref{fig:ItalianStat}c, shows that the mean number of errors recorded during the SR test was 3.58 for participants with SLD$_{i}$ and 3.45 for those in the CG$_{i}$. Finally, the mean scores, Figure~\ref{fig:ItalianStat}d, for RSES were 23.9 for the SLD$_{i}$ group and 24.5 for the CG$_{i}$.\vspace{-6pt}

Table~\ref{tab: ItalianTab} presents the results of statistical tests for the Italian sample. The \emph{t}-test results indicate statistically significant differences in the time taken to complete the SR and RSES tasks (\emph{p} < $0.001$ for SR time and \emph{p} = $0.003$ for RSES time), {and the power analysis showed high sensitivity to the large effects (power $= 0.999$ for SR time and power $= 0.932$ for RSES time)}. This suggests that the SLD group took significantly different amounts of time to complete both tasks compared to the control group. Crucially, the Mann--Whitney U test results show no significant differences for SR errors (\emph{p} = $0.584$) and RSES scores (\emph{p} = $0.531$). This indicates that there was no significant difference in the number of errors made in the SR test or in the self-esteem scores between the two groups.

\begin{table}[H] 
    \caption{Results of \emph{t}-test and Mann--Whitney test for the Italian sample.\label{tab: ItalianTab}}
    \begin{tabularx}{\textwidth}{CCC}
        \toprule
        \textbf{}	& \textbf{SR}	& \textbf{RSES}\\
        \midrule
        \emph{p}-value (power)		& <0.001 (0.999)	&  0.003 (0.932)\\
        
        \emph{p}-value		& $0.584  $	& $0.531$\\
        \bottomrule
    \end{tabularx}
\end{table}\clearpage 

\subsubsection{Classifier Performance}

\begin{figure}[H]
    \subfloat[][\centering]{\includegraphics[width=1\textwidth]{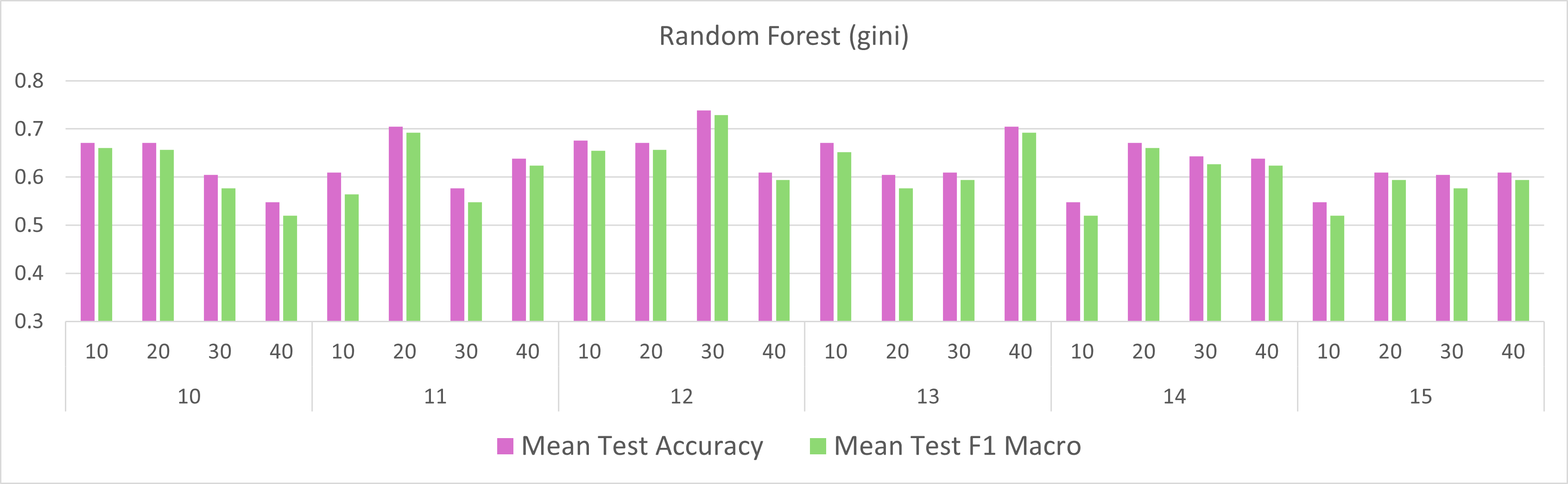}\label{subfig:RF_g_ITA}}\\
    \subfloat[][\centering]{\includegraphics[width=1\textwidth]{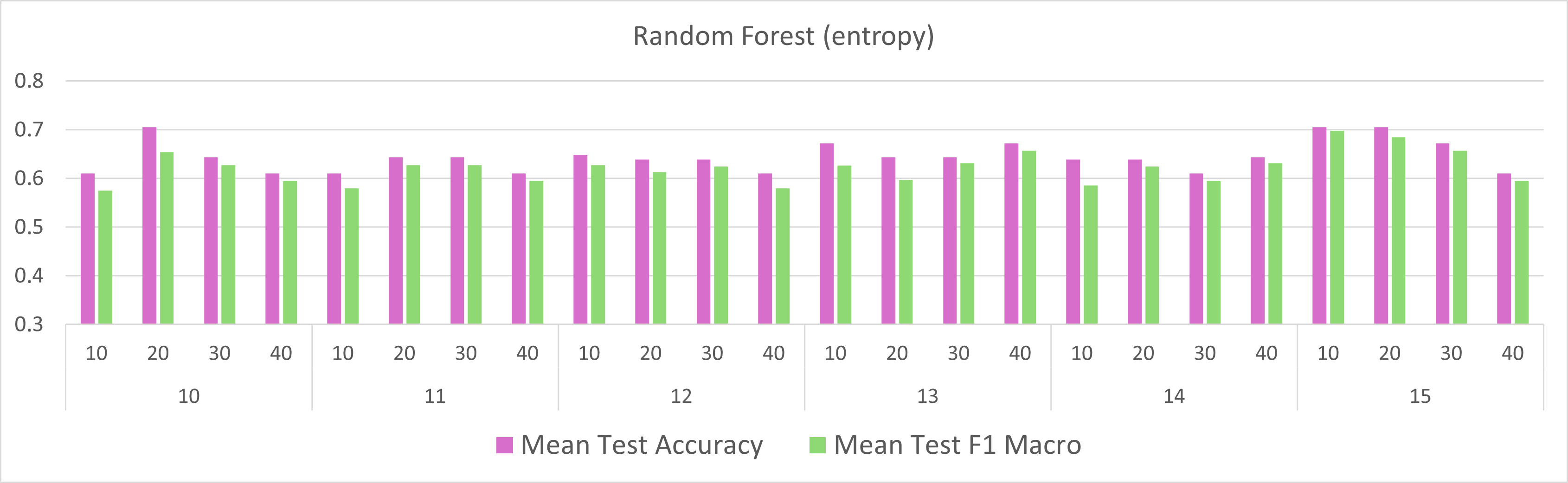}\label{subfig:RF_e_ITA}}\\
    \subfloat[][\centering]{\includegraphics[width=0.485\textwidth]{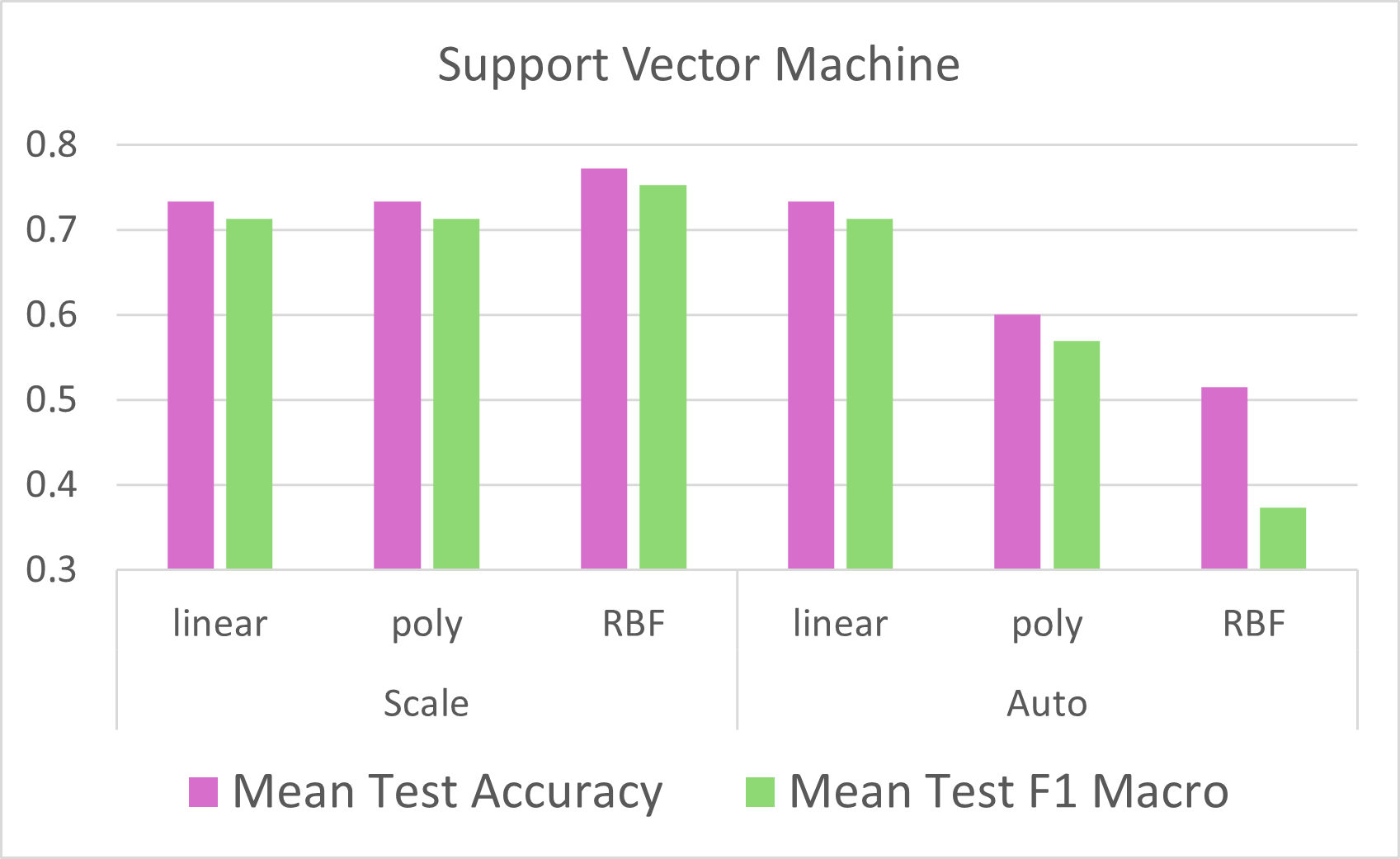}\label{subfig:SVM_ITA}}\quad
    \subfloat[][\centering]{\includegraphics[width=0.485\textwidth]{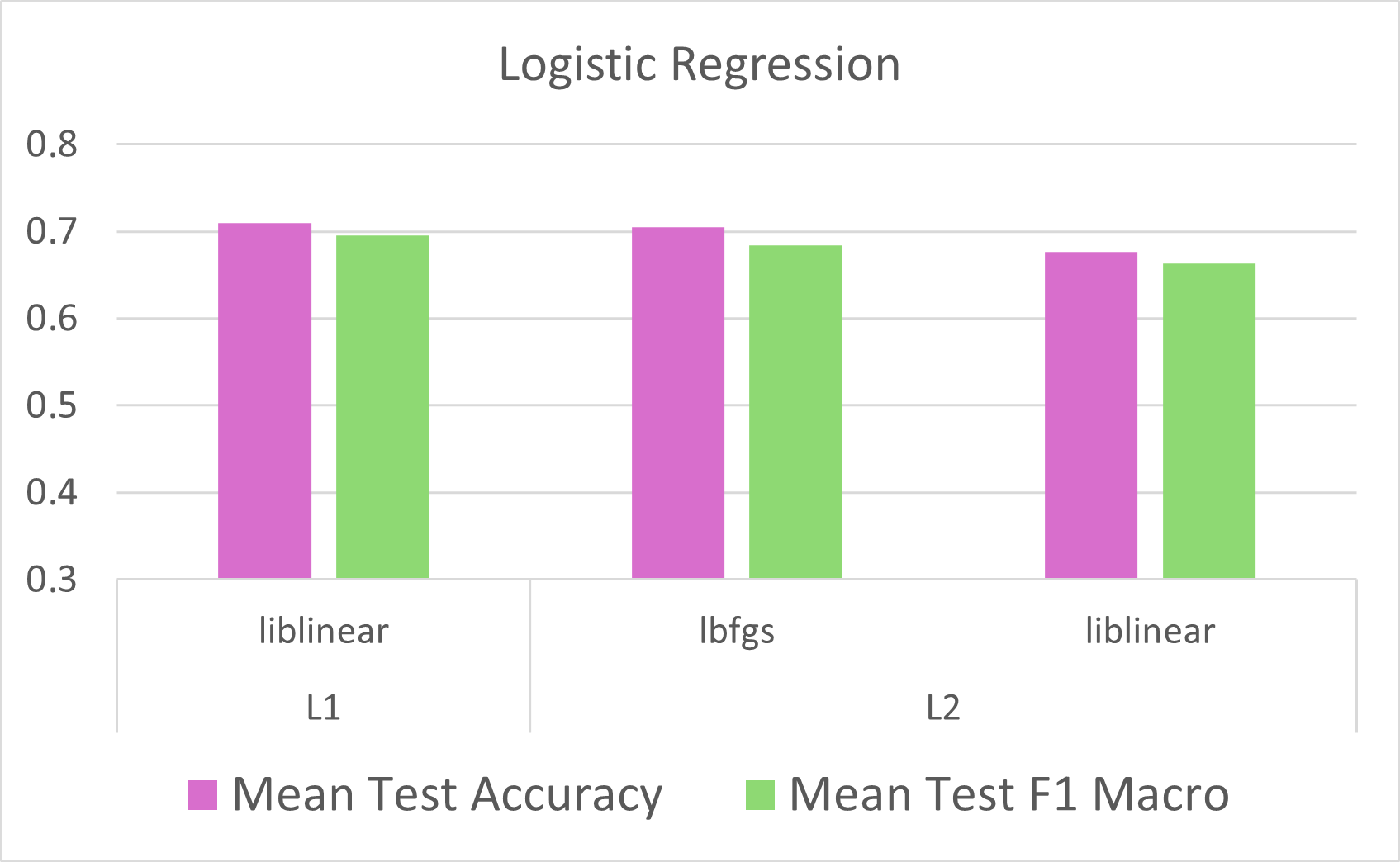}\label{subfig:LR_ITA}} \\
	\subfloat[][\centering]{\includegraphics[width=0.485\textwidth]{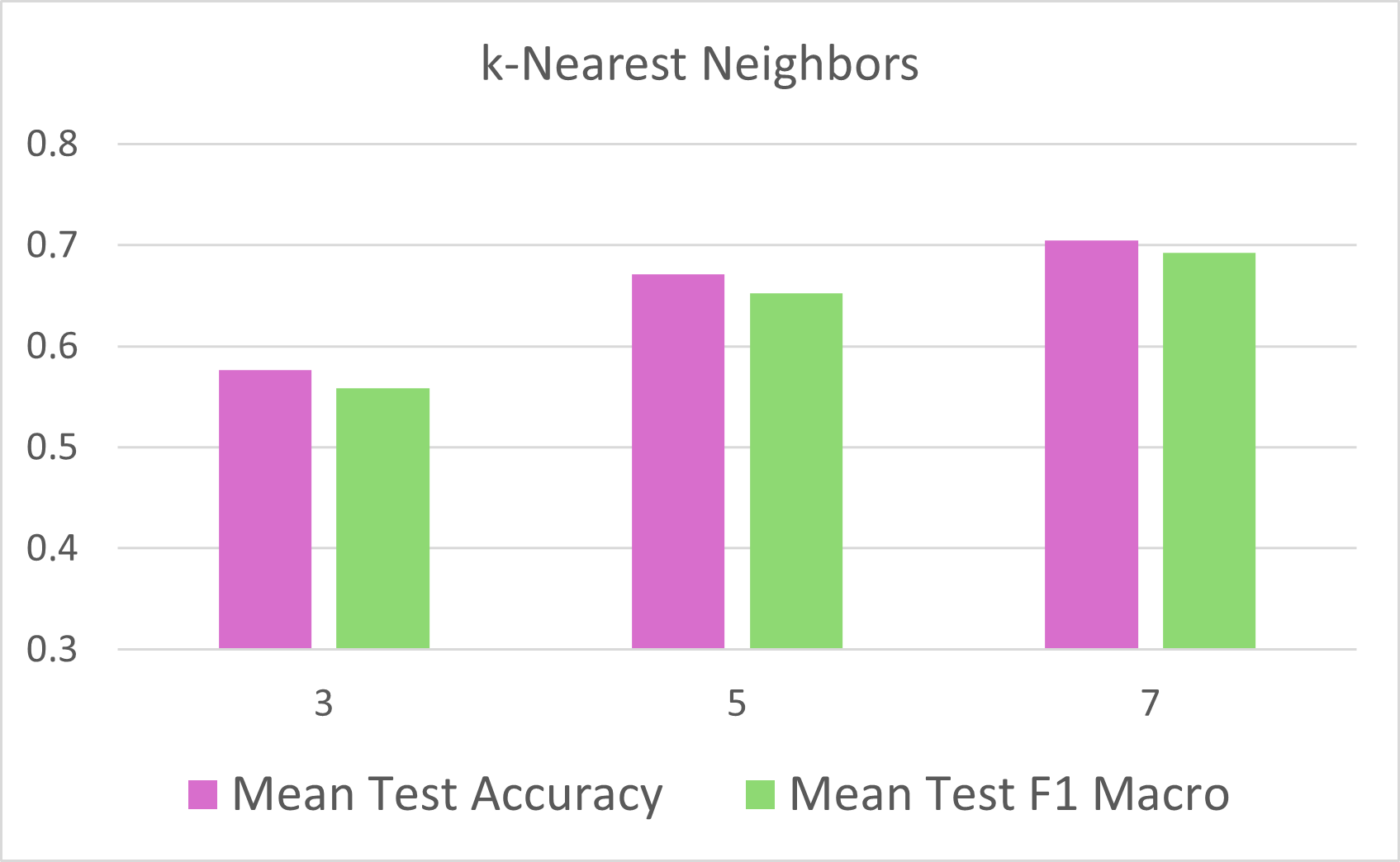}\label{subfig:KNN_ITA}}\quad
	\subfloat[][\centering]{\includegraphics[width=0.485\textwidth]{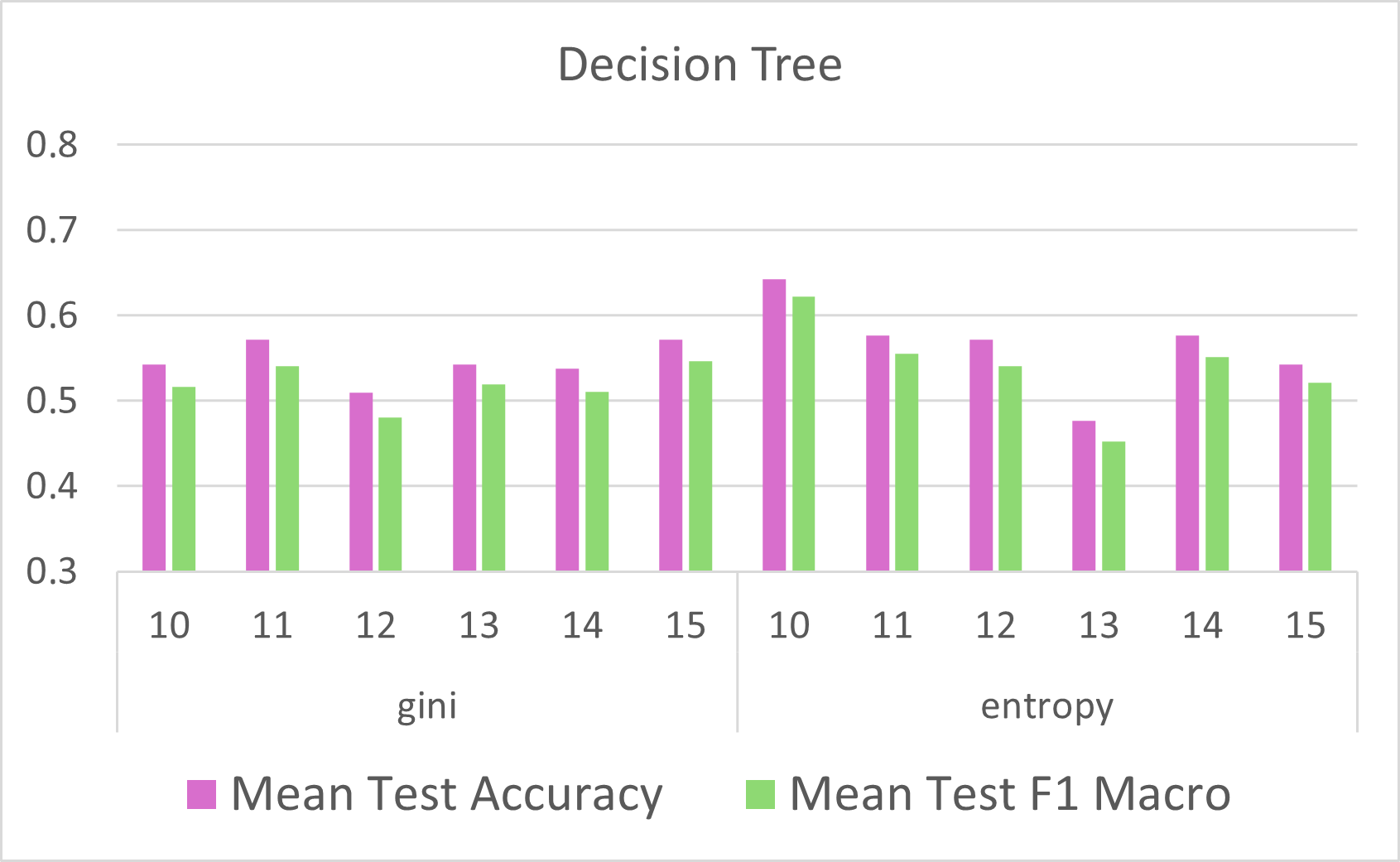}\label{subfig:DT_ITA}} \\
    \caption{{Performance} of the ML algorithms for the Italian group. {(\textbf{a}) RF with gini criterion; (\textbf{b}) RF with entropy criterion; (\textbf{c}) SVM; (\textbf{d}) LR; (\textbf{e}) k-NN; (\textbf{f}) DT.}} 
    \label{fig:ItalianClass}
\end{figure}

The results of the the training session for the Italian user group will now be presented. The SVM demonstrated a strong capacity to effectively model the data, Figure~\ref{fig:ItalianClass}c. Notably, the SVM with the RBF kernel and ''scale'' gamma achieved the highest performance, with an accuracy of 77.1\% and an F1-score of 75.3\%. This result indicates that, among the algorithms tested, the SVM was particularly adept at capturing the underlying patterns in the user data. RF was also a competitive algorithm within the Italian group, Figure~\ref{fig:ItalianClass}a,b. Its performance, however, was more variable and sensitive to the chosen parameters, such as the maximum depth of the trees and the number of estimators. The best results observed for RF were around 70.5\% for accuracy and 69.7\% for the F1-score. For instance, when using the gini criterion and a max depth of 11, the accuracy reached 70.5\% with 20 estimators. LR provided a more stable performance baseline for the Italian group, Figure~\ref{fig:ItalianClass}d. It consistently delivered accuracy scores around 70\% and F1-scores around 68\% across the different parameter settings. As an example, the L1 penalty combined with the Liblinear solver yielded an accuracy of 71.0\% and an F1-score of 70.0\%. In contrast, the DT classifier showed a performance range with accuracies generally between 50\% and 60\%, Figure~\ref{fig:ItalianClass}f. The entropy criterion tended to produce slightly better results compared with the gini criterion, but overall, the DT's predictive power was lower than that of SVM, RF, or LR in this group.  Finally, k-NN exhibited a trend of improving performance as the number of neighbors increased, Figure~\ref{fig:ItalianClass}e. The highest accuracy achieved by k-NN was approximately 70\% when using seven neighbors, suggesting that considering a larger neighborhood enhanced the model's ability to generalize.

Therefore, the best configuration (SVM with the RBF kernel and adjusting the gamma based on the range of input values) was used on the test set, achieving 87.5\% for accuracy and 85.7\% for F1-score.

\subsection{Spanish Group}
\subsubsection{Differences Between Groups}
\begin{figure}[H]
	\subfloat[][\centering]{\includegraphics[width=0.48\textwidth]{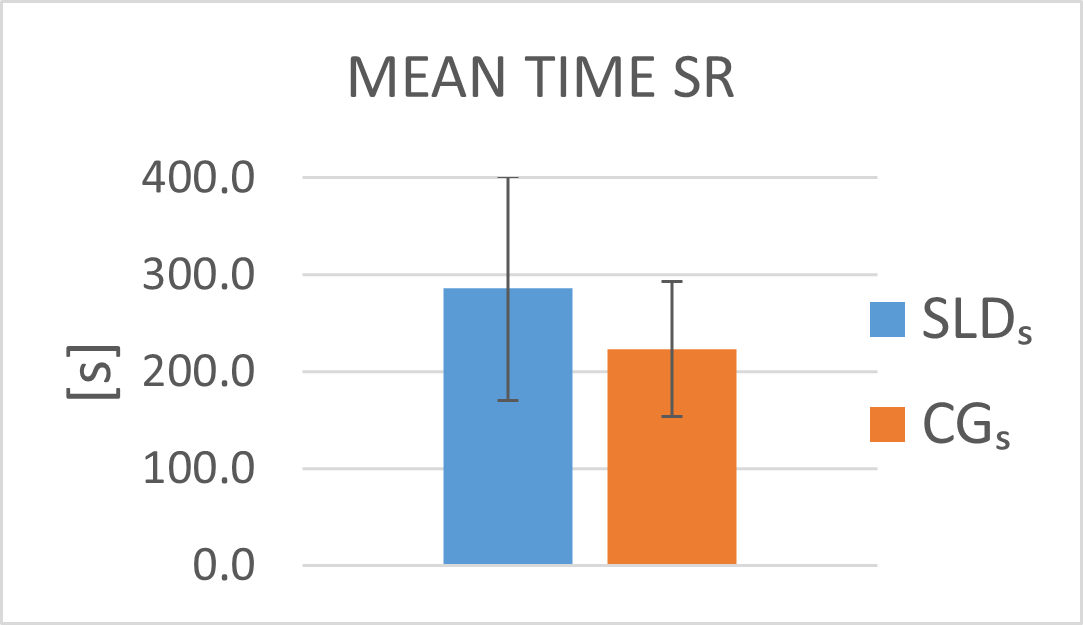}\label{subfig:MTSR_SPA}} \quad
	\subfloat[][\centering]{\includegraphics[width=0.48\textwidth]{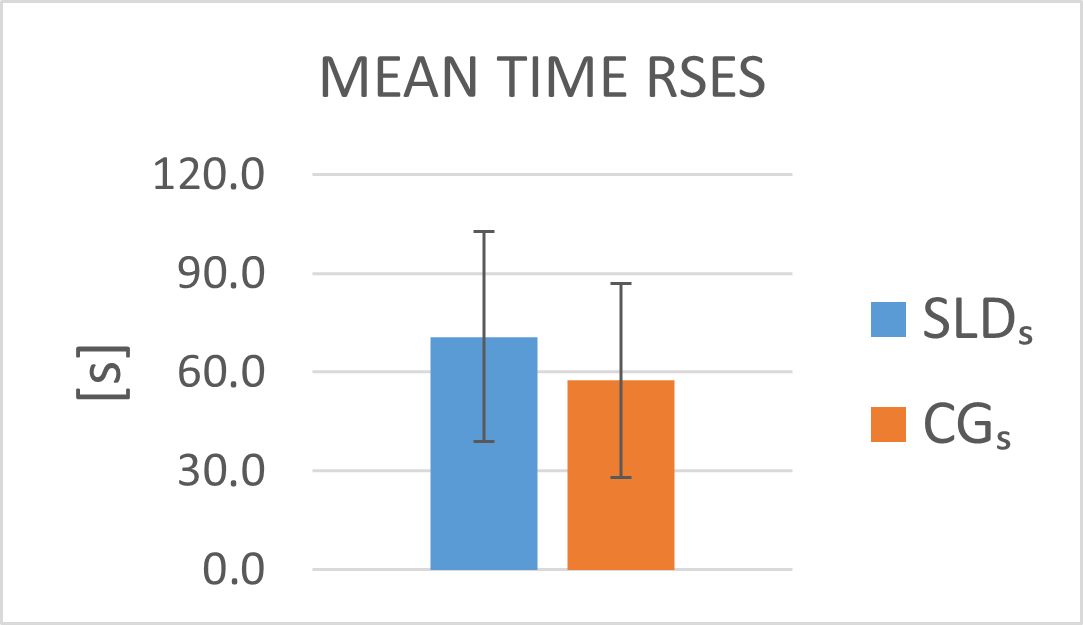}\label{subfig:MTRSES_SPA}}\\
    \subfloat[][\centering]{\includegraphics[width=0.48\textwidth]{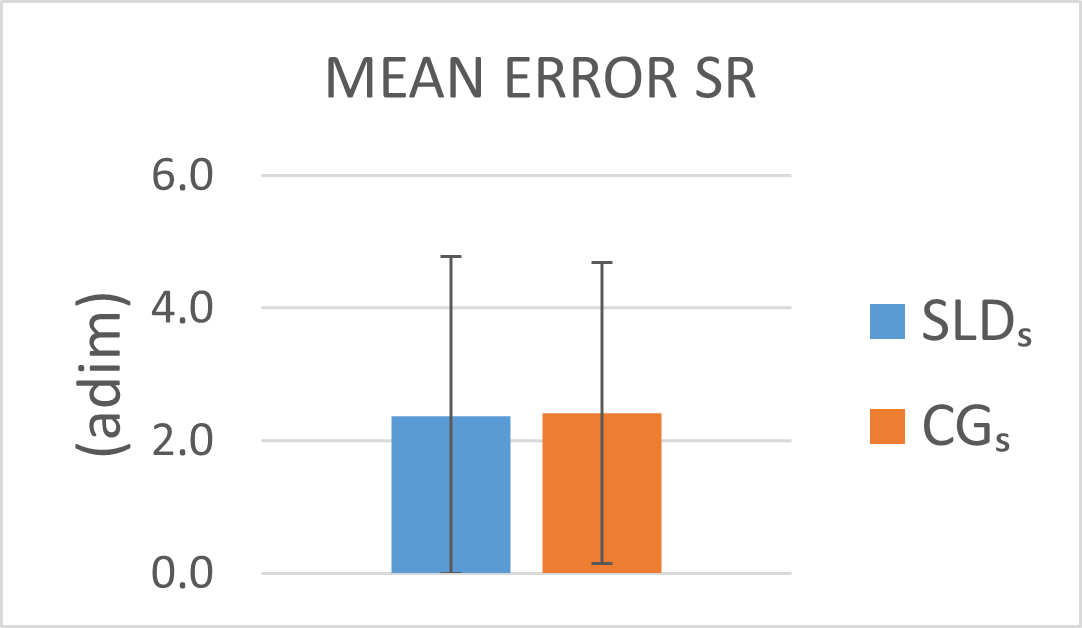}\label{subfig:MESR_SPA}} \quad
    \subfloat[][\centering]{\includegraphics[width=0.48\textwidth]{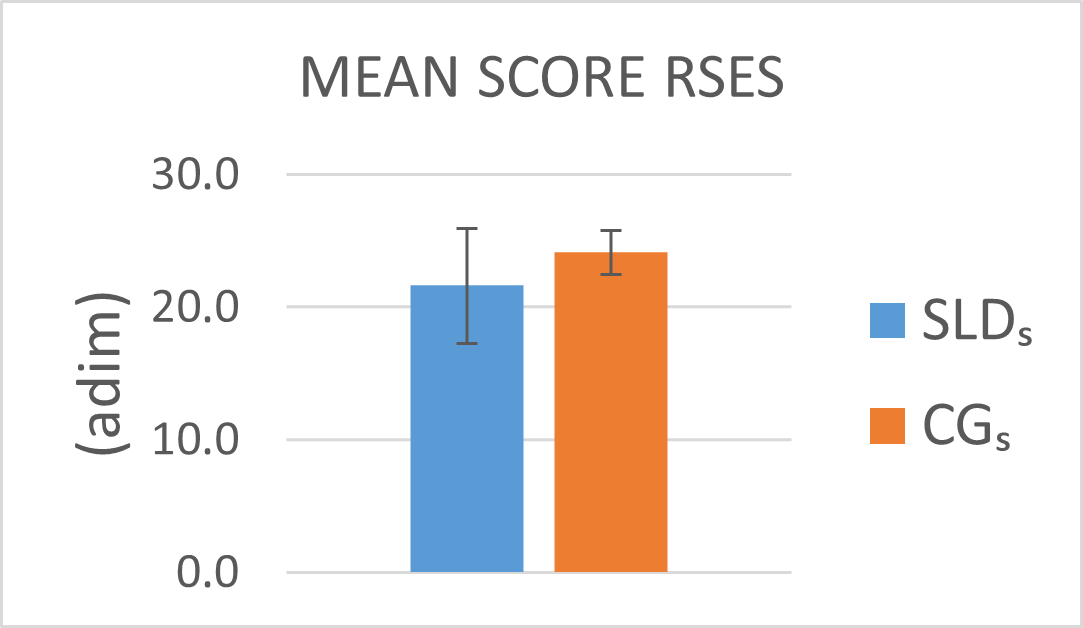}\label{subfig:MSRSES_SPA}}\\
    \caption{{Performance} of the Spanish group: (\textbf{a}) average time to perform SR test; (\textbf{b}) average time to perform RSES test; (\textbf{c}) average errors made during SR test; (\textbf{d}) average of the scores obtained in the RSES test.}
    \label{fig:SpanishData}
\end{figure}

In Figure~\ref{fig:SpanishData}, as previously, from left to right and from top to bottom, the first graphic \mbox{Figure~\ref{fig:SpanishData}a} shows the mean time  needed to finish the SR test, with 264 s for the SLD$_{s}$ group and 196 s for the CG$_{s}$. The data results related to  the mean time required to complete the RSES test, Figure~\ref{fig:SpanishData}b, illustrate that the SLD$_{s}$ participants spent 70.7 s, with the CG$_{s}$  completing it in 57.5 s. The third graphic, Figure~\ref{fig:SpanishData}c, is related to the mean errors recorded during the SR test, showing a mean of 2.95 for the SLD$_{s}$ group and 3.40 for the CG$_{s}$. Lastly, the mean scores recorded in the RSES, Figure~\ref{fig:SpanishData}d, were 21.6 for the SLD$_{s}$ group and 24.1 for the CG$_{s}$.

 Table~\ref{tab: SpanishTab} shows the statistical test results for the Spanish sample. The \emph{t}-test results revealed no statistically significant differences in task completion time between groups for either SR (\emph{p} = $0.063$) or RSES (\emph{p} = $0.174$), the {power analysis showed a moderate sensitivity to large effects for SR time (power $= 0.522$) and a low sensitivity for RSES time (\mbox{power $= 0.264$})}.  Consistently, the Mann--Whitney U test also showed no significant differences for SR errors (\emph{p} = $0.696 $) and RSES scores (\emph{p} = $0.069$).

 \begin{table}[H] 
    \caption{Results of \emph{t}-test and Mann--Whitney test for the Spanish sample.\label{tab: SpanishTab}}
    \begin{tabularx}{\textwidth}{CCC}
        \toprule
        \textbf{}	& \textbf{SR}	& \textbf{RSES}\\
        \midrule
        \emph{p}-value (power)		& $0.063$ ($0.522$) 	& $0.174$ ($0.264$)\\
        \emph{p}-value		& $0.696$	& $0.069 $\\
        \bottomrule
    \end{tabularx}
\end{table}

\subsubsection{Classifier Performance}

\begin{figure}[H]
    \centering
    \subfloat[][\centering]{\includegraphics[width=1\textwidth]{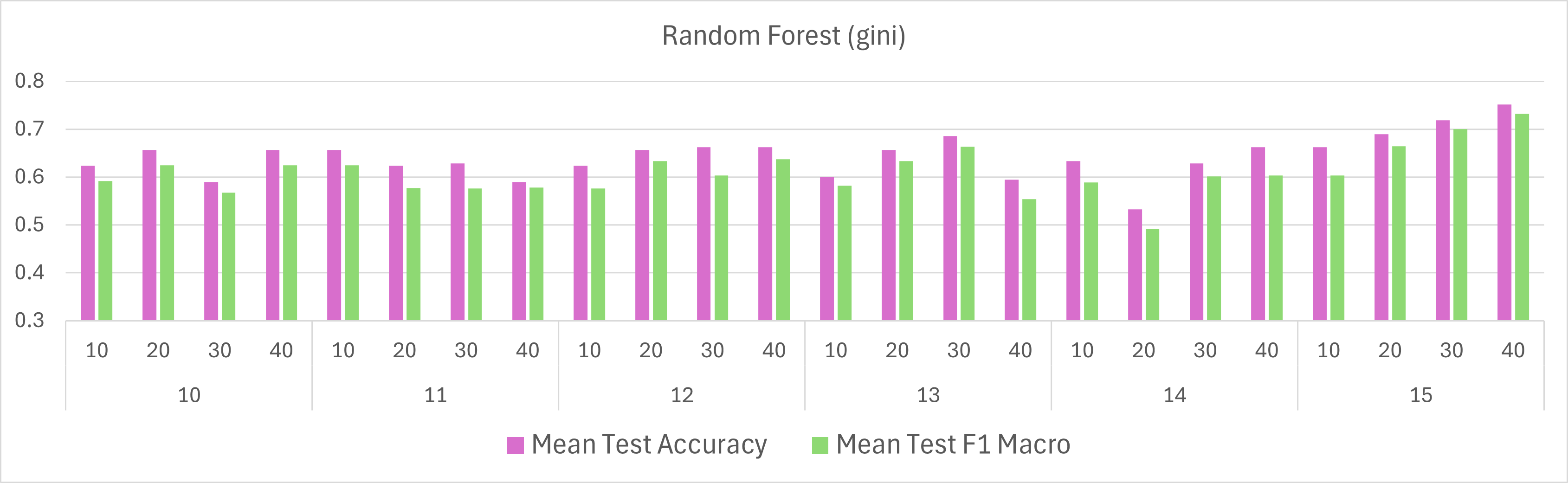}\label{subfig:RF_g_SPA}}\\
    \subfloat[][\centering]{\includegraphics[width=1\textwidth]{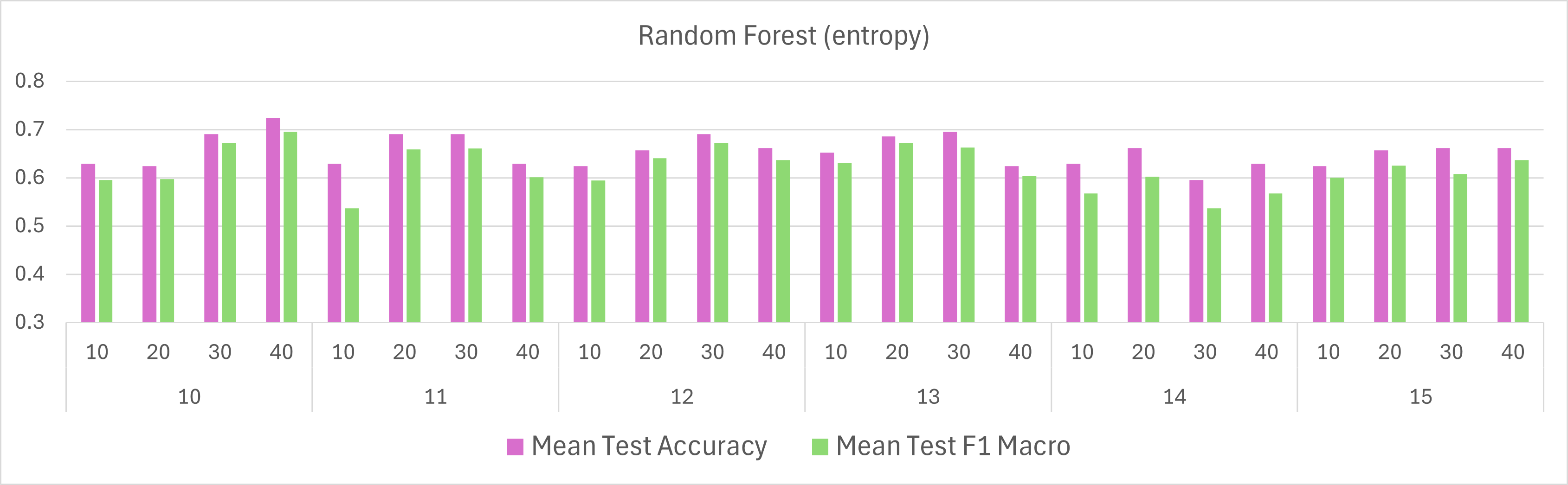}\label{subfig:RF_e_SPA}}\\
    \subfloat[][\centering]{\includegraphics[width=0.485\textwidth]{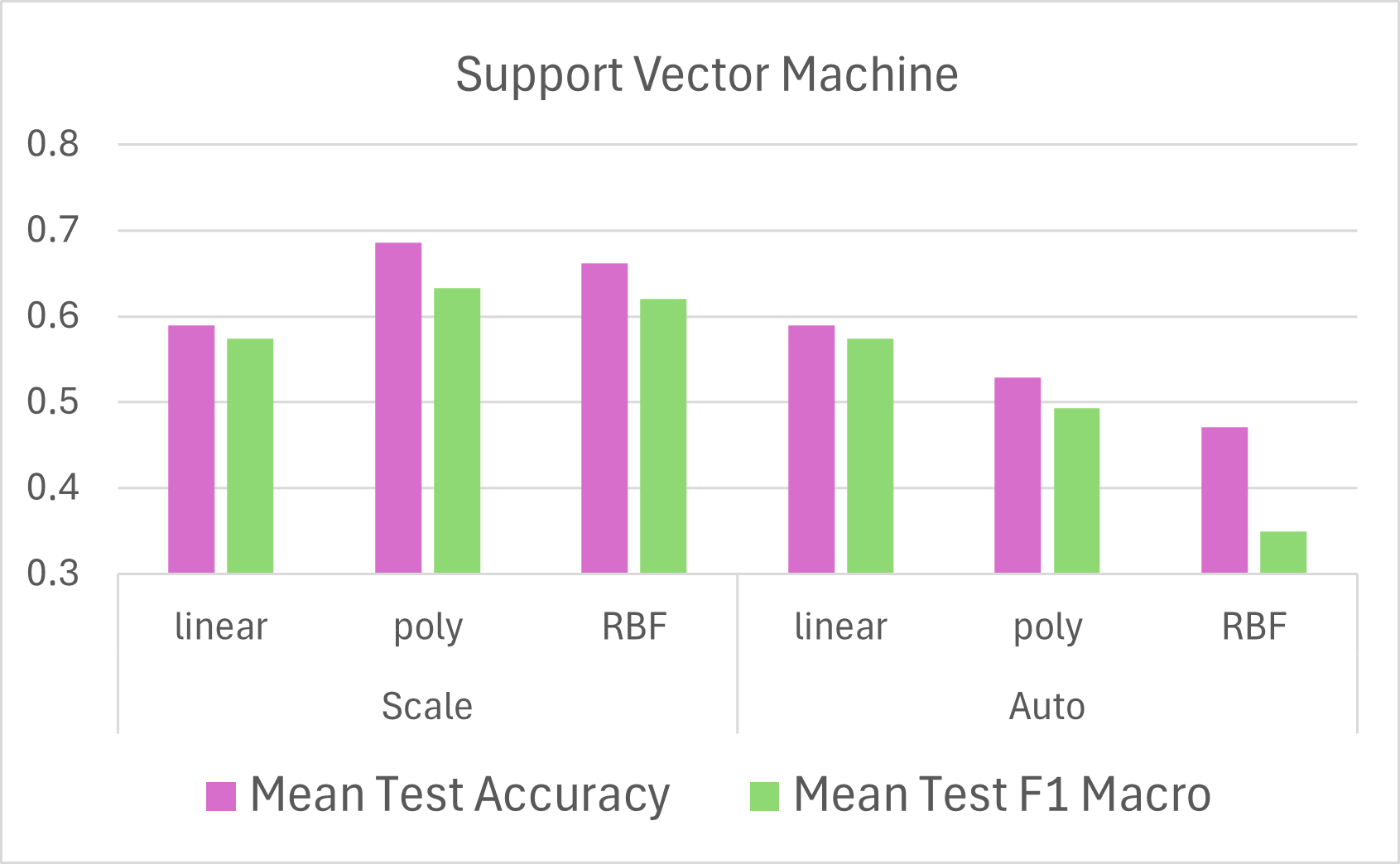}\label{subfig:SVM_SPA}}\quad
    \subfloat[][\centering]{\includegraphics[width=0.485\textwidth]{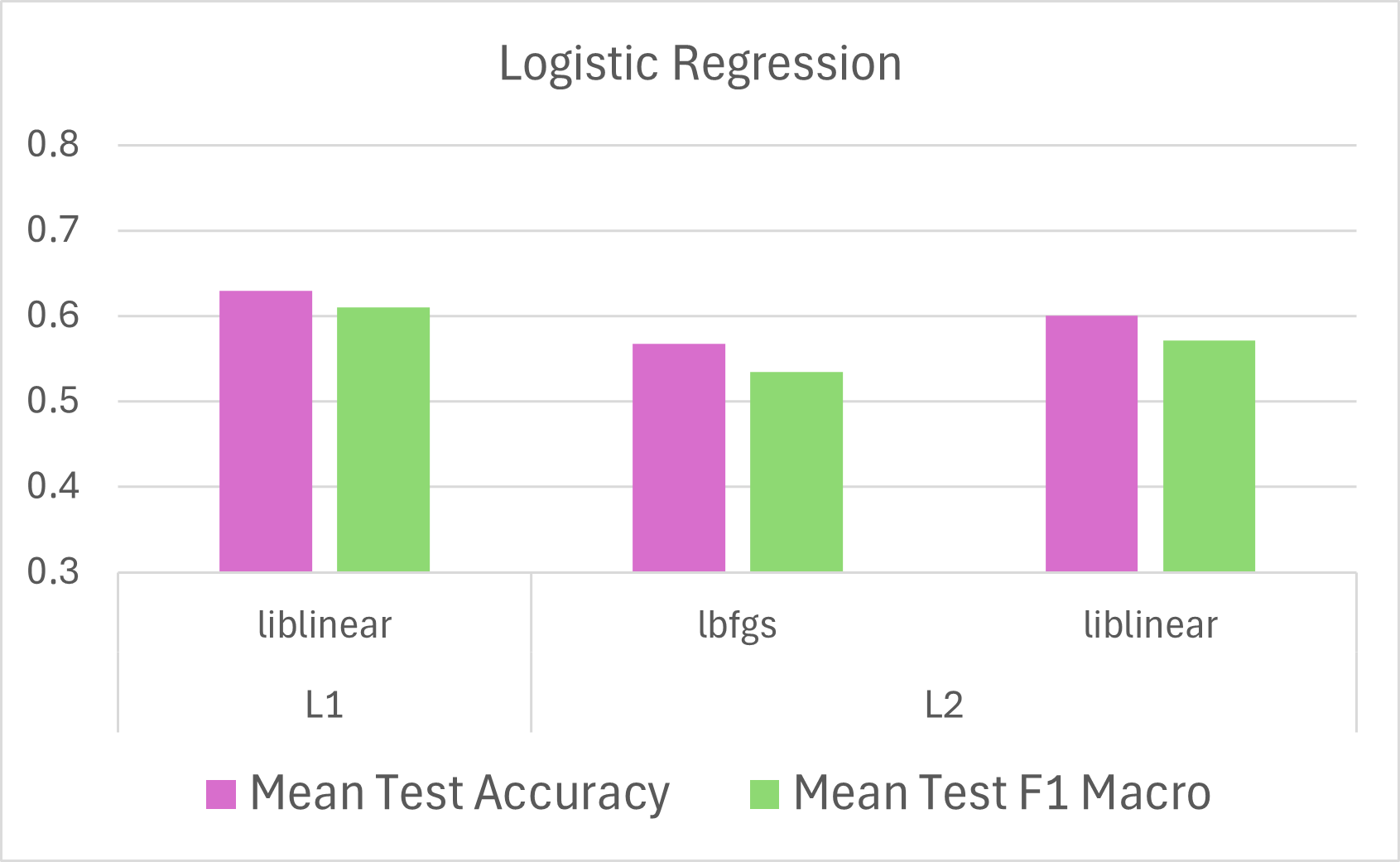}\label{subfig:LR_SPA}} \\
	\subfloat[][\centering]{\includegraphics[width=0.485\textwidth]{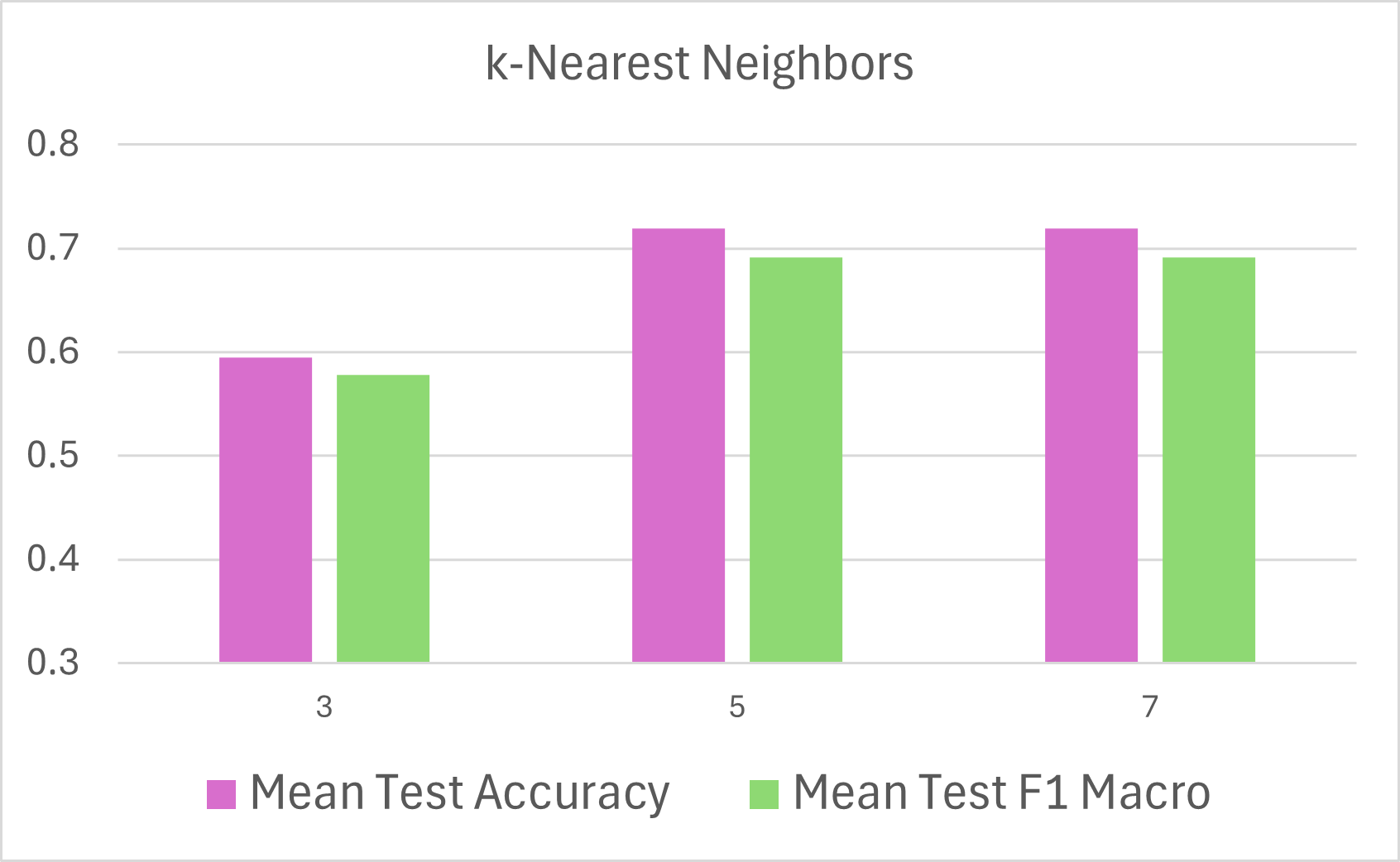}\label{subfig:KNN_SPA}}\quad
	\subfloat[][\centering]{\includegraphics[width=0.485\textwidth]{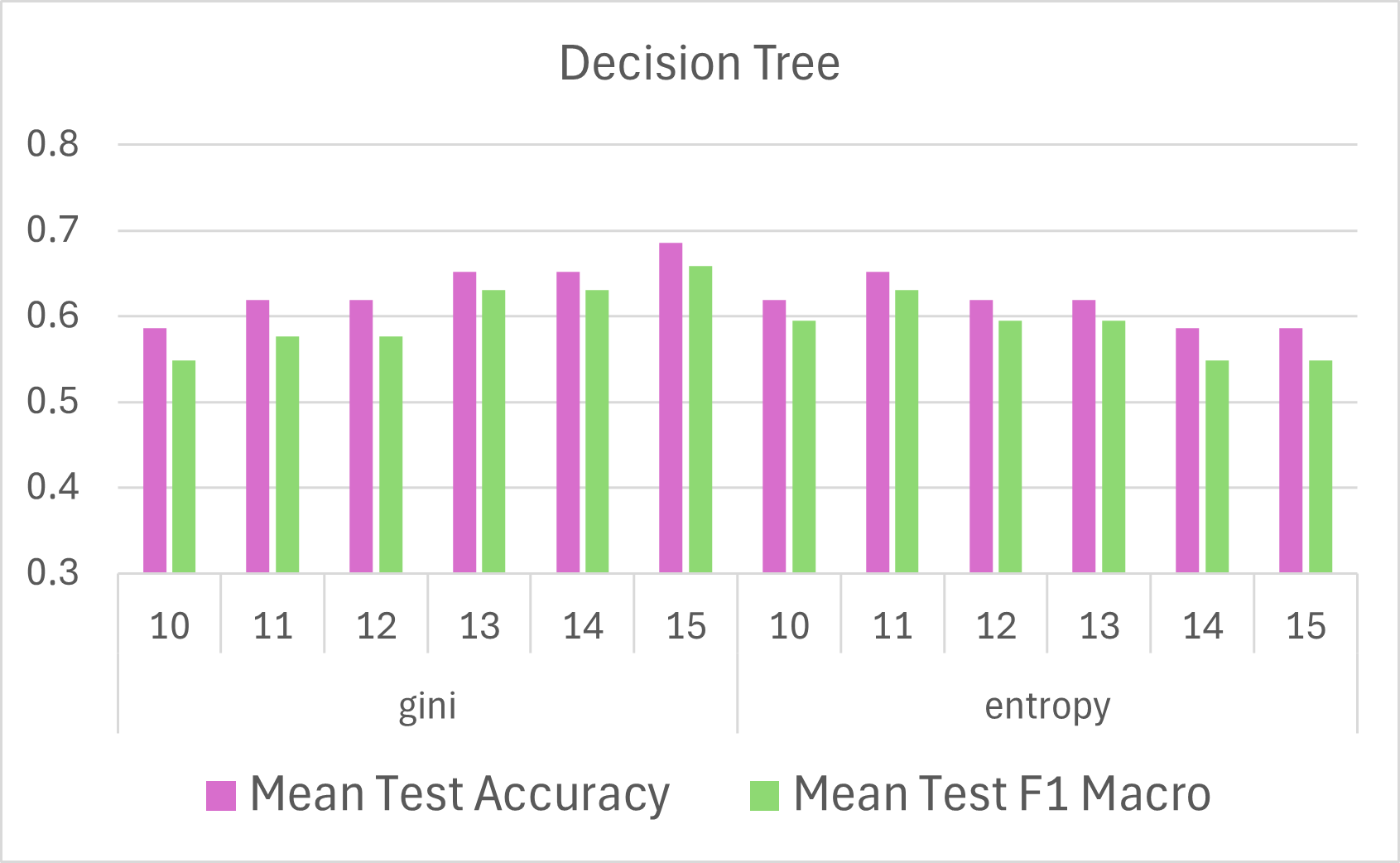}\label{subfig:DT_SPA}} \\
    \caption{{Performance}  of the ML algorithms on the Spanish group.  {(\textbf{a}) RF with gini criterion; (\textbf{b}) RF with entropy criterion; (\textbf{c}) SVM; (\textbf{d}) LR; (\textbf{e}) k-NN; (\textbf{f}) DT.}}
    \label{fig:SpanishClass}
\end{figure}

Moving on to the ML training session for the Spanish user group, the RF algorithm proved to be particularly effective, Figure~\ref{fig:SpanishClass}a,b. It achieved the highest accuracy, reaching up to 75.2\% and an F1-score of 73.2\%. This peak performance was observed with the gini criterion, a max depth of 15, and 40 estimators, demonstrating RF's capacity to model the complexities of the data. k-NN also performed reasonably well for this group of data, Figure~\ref{fig:SpanishClass}e. It obtained an accuracy of 71.9\% and an F1-score of 69.1\% when using either five or seven neighbors. This indicates that k-NN can be a viable alternative, though it did not quite reach the peak performance of RF in this context. The SVM showed a more varied performance in the Spanish group, compared to its performance in the Italian group, Figure~\ref{fig:SpanishClass}c. While the polynomial kernel achieved an accuracy of 68.6\% and an F1-score of 63.3\%, SVM did not outperform RF. The effectiveness of SVM appeared to be more sensitive to the choice of kernel and parameters in the Spanish data. LR generally exhibited accuracy in the lower 60\% range Figure~\ref{fig:SpanishClass}d. This suggests that while LR could provide a stable baseline, it may not have fully captured the nuances present in the user data. Similarly to the first group, the DT classifier's performance for the Spanish group was generally around 60\% accuracy,  Figure~\ref{fig:SpanishClass}f. This consistency across both groups indicates that DT, while simple, may lack the predictive power of more complex algorithms like RF or SVM.   

Again, the best configuration was used on the test set. As just mentioned, this time the best performance in training was achieved by the RF classifier with the gini criterion, a max depth of 15, and 40 estimators, achieving 66.6\% for both accuracy and F1-score on the test set.

\subsection{Pooled Results}
\subsubsection{Differences Between Groups}
\begin{figure}[H]
	\subfloat[][\centering]{\includegraphics[width=0.45\textwidth]{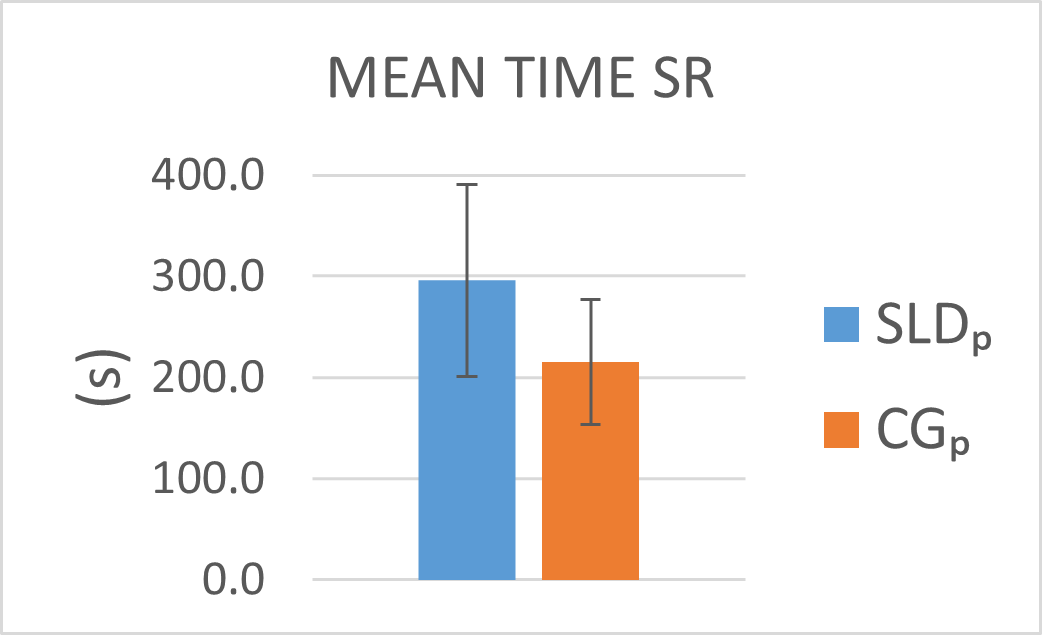}\label{subfig:MTSR_ALL}} \quad
	\subfloat[][\centering]{\includegraphics[width=0.45\textwidth]{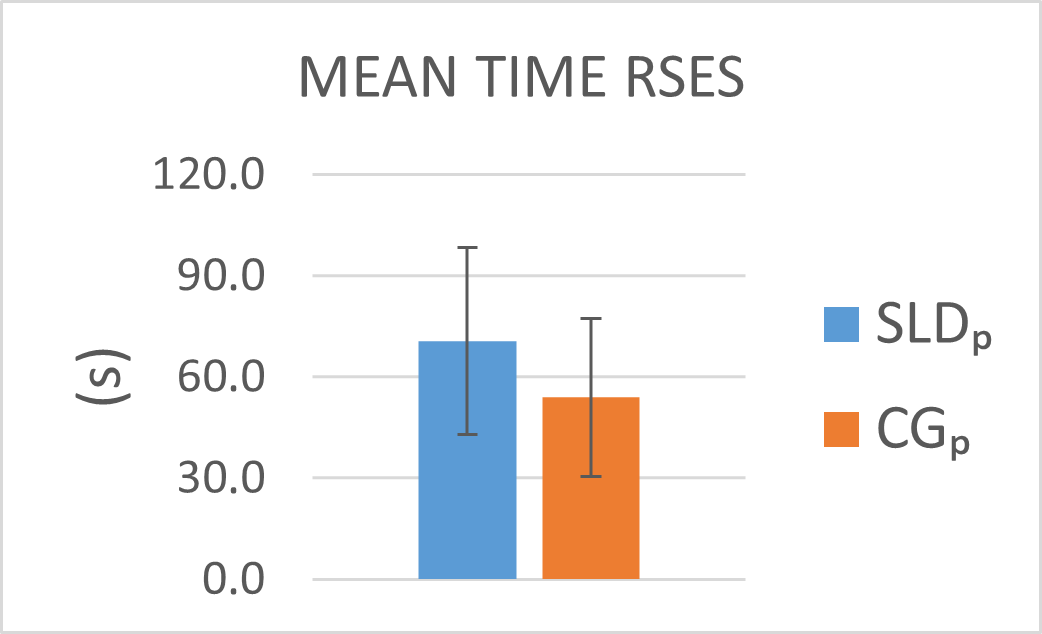}\label{subfig:MTRSES_ALL}}\\
    \subfloat[][\centering]{\includegraphics[width=0.45\textwidth]{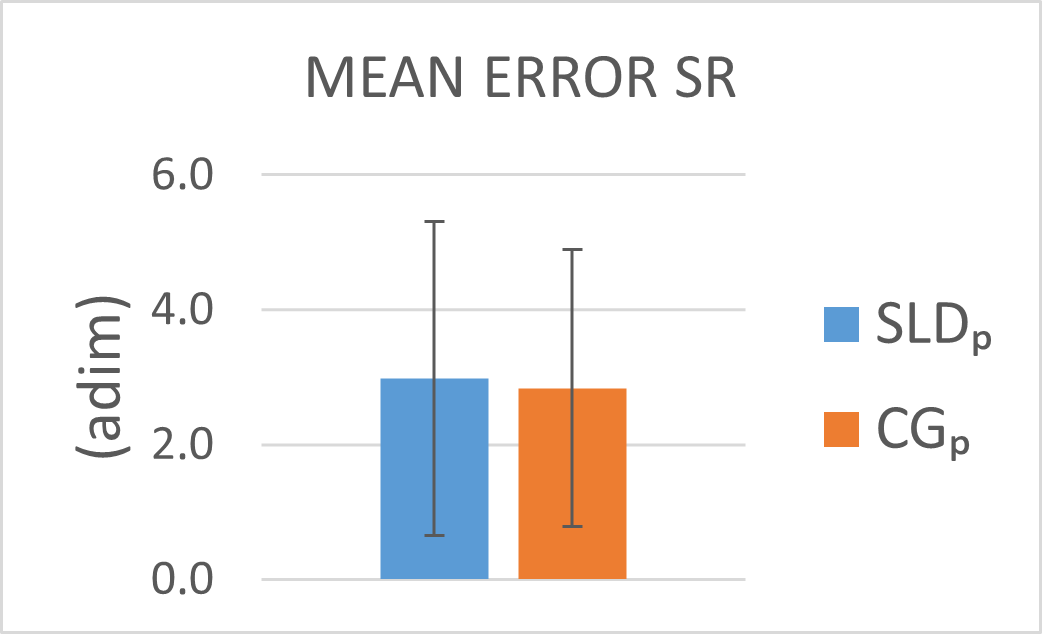}\label{subfig:MESR_ALL}} \quad
    \subfloat[][\centering]{\includegraphics[width=0.45\textwidth]{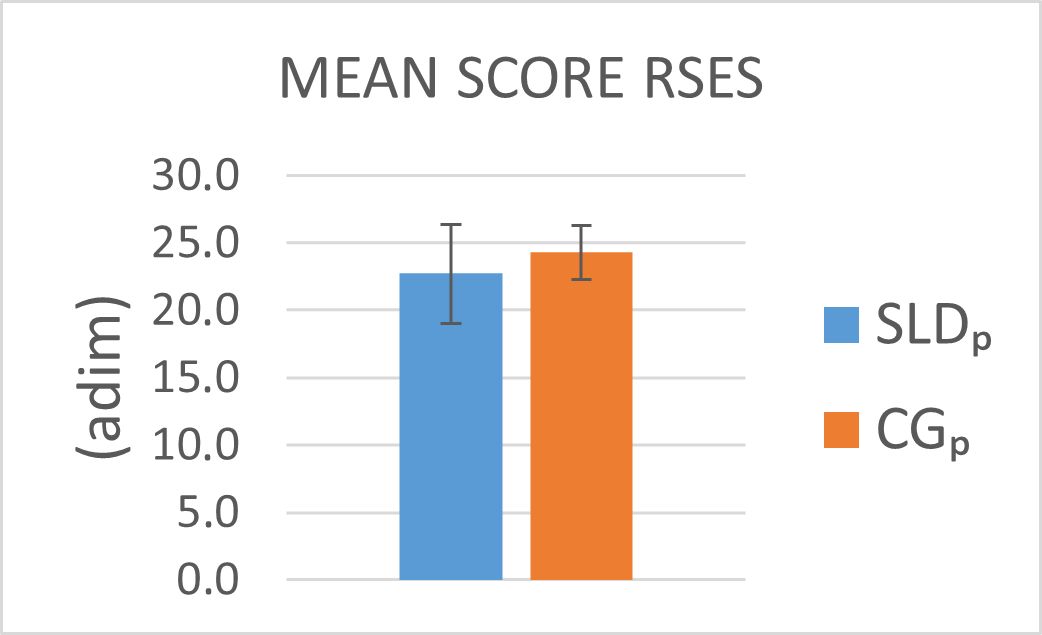}\label{subfig:MSRSES_ALL}}\\
    \caption{{Performance} of the Pooled group: (\textbf{a}) average time to perform SR test; (\textbf{b}) average time to perform RSES test; (\textbf{c}) average errors made during SR test; (\textbf{d}) average of the scores obtained in the RSES test.}
    \label{fig:CombinedData}
\end{figure}

\textls[-15]{In {Figure}~\ref{fig:CombinedData}, the graphics related to the results using the entire dataset are presented. The graphic related to the mean time taken to finish, Figure~\ref{fig:CombinedData}a, shows that the results for the SR test were 284 s for the SLD$_{m}$ group and 198 s for the CG$_{m}$. Concerning the mean time required to complete the RSES test, Figure~\ref{fig:CombinedData}b, it is possible to see that the SLD$_{m}$ participants required 70.5 s and the CG$_{m}$ needed 53.9 s. The diagram of the mean errors of the SR test, \mbox{Figure~\ref{fig:CombinedData}c}, illustrate a mean of 3.25 for  SLD$_{m}$ participants and 3.43 for the CG$_{m}$. Finally, the mean scores achieved in the RSES Figure~\ref{fig:CombinedData}d were 22.7 for the SLD$_{m}$ group and 24.3 for the CG$_{m}$.\vspace{-6pt}}

\textls[-15]{Table~\ref{tab: PoolTab} presents the statistical test results for the entire sample. The \emph{t}-test results indicate statistically significant differences in task completion time for SR (\emph{p} {<}  0.001) and RSES (\mbox{\emph{p} = $0.005$}). {The power analysis also shows a high sensitivity to the large effects for SR time (power $= 0.994$)  and for RSES time (power $= 0.814$)}.  However, the Mann--Whitney U test results showed no significant differences for SR errors (\emph{p} = $0.952$) and RSES scores (\emph{p} = $0.853$).}
 
 \begin{table}[H] 
    \caption{Results of \emph{t}-test and Mann--Whitney test for the entire sample.\label{tab: PoolTab}}
    \begin{tabularx}{\textwidth}{CCC}
        \toprule
        \textbf{}	& \textbf{SR}	& \textbf{RSES}\\
        \midrule
        \emph{p}-value (power)		& <0.001 ($0.994$)	& $0.005$ ($0.814$)\\
        \emph{p}-value		& $0.952$	& $0.853$\\
        \bottomrule
    \end{tabularx}
\end{table}

\subsubsection{Classifier Performance}

\begin{figure}[H]
    
    \subfloat[][\centering]{\includegraphics[width=0.975\textwidth]{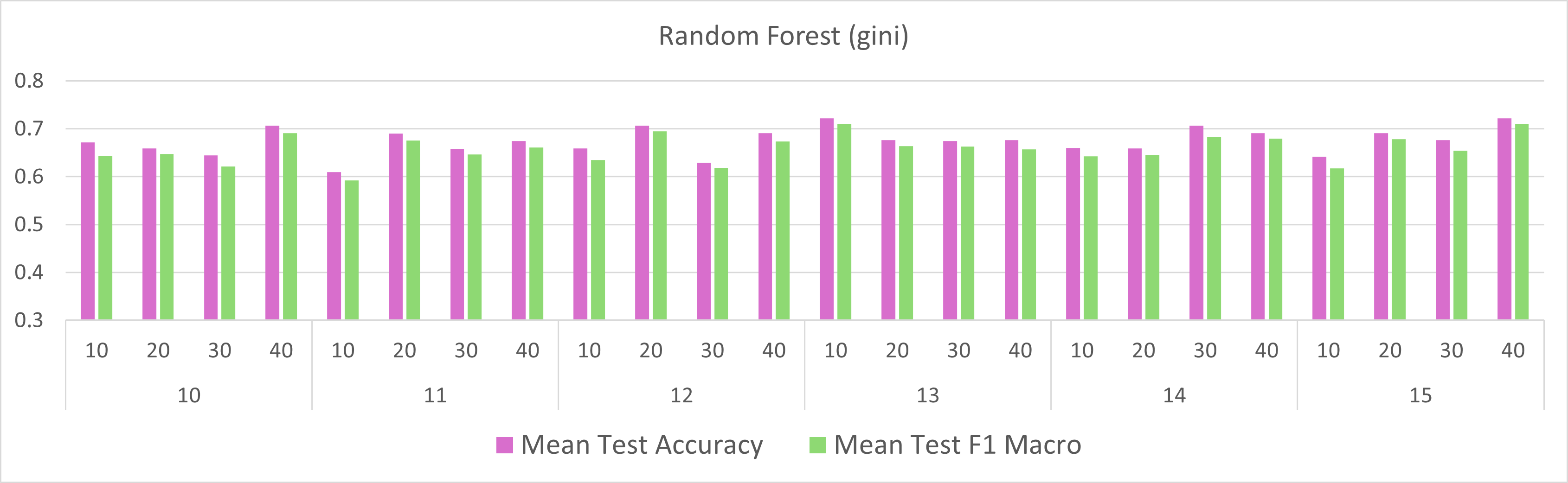}\label{subfig:RF_g_ALL}}\\
    \subfloat[][\centering]{\includegraphics[width=0.975\textwidth]{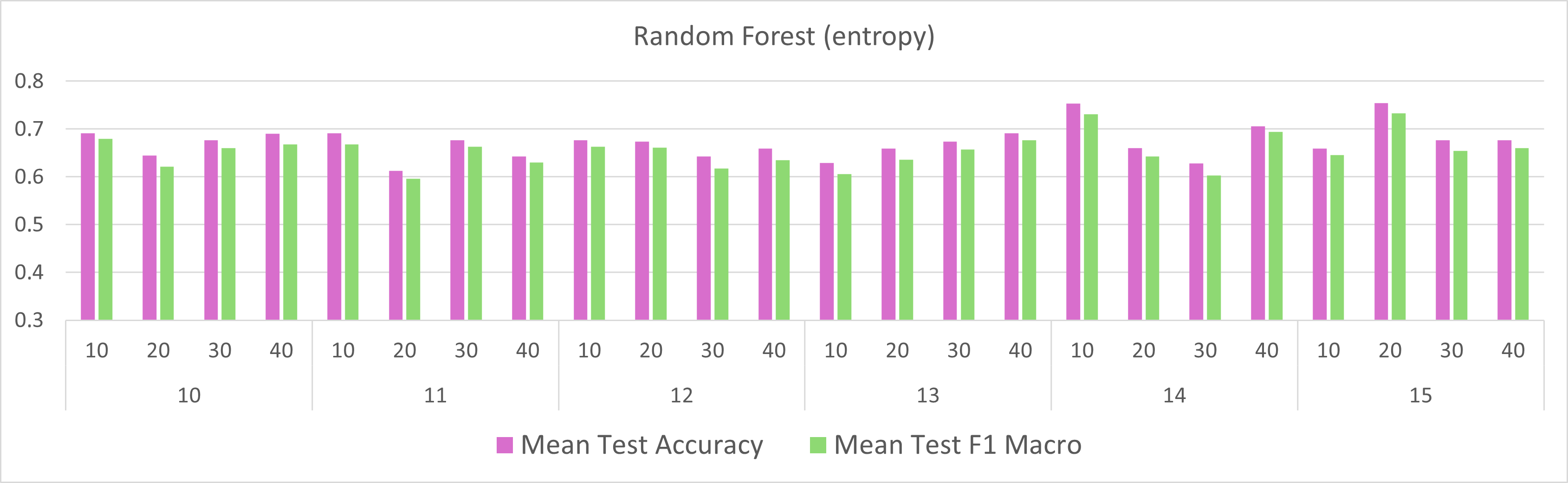}\label{subfig:RF_e_ALL}}\\
    \subfloat[][\centering]{\includegraphics[width=0.47\textwidth]{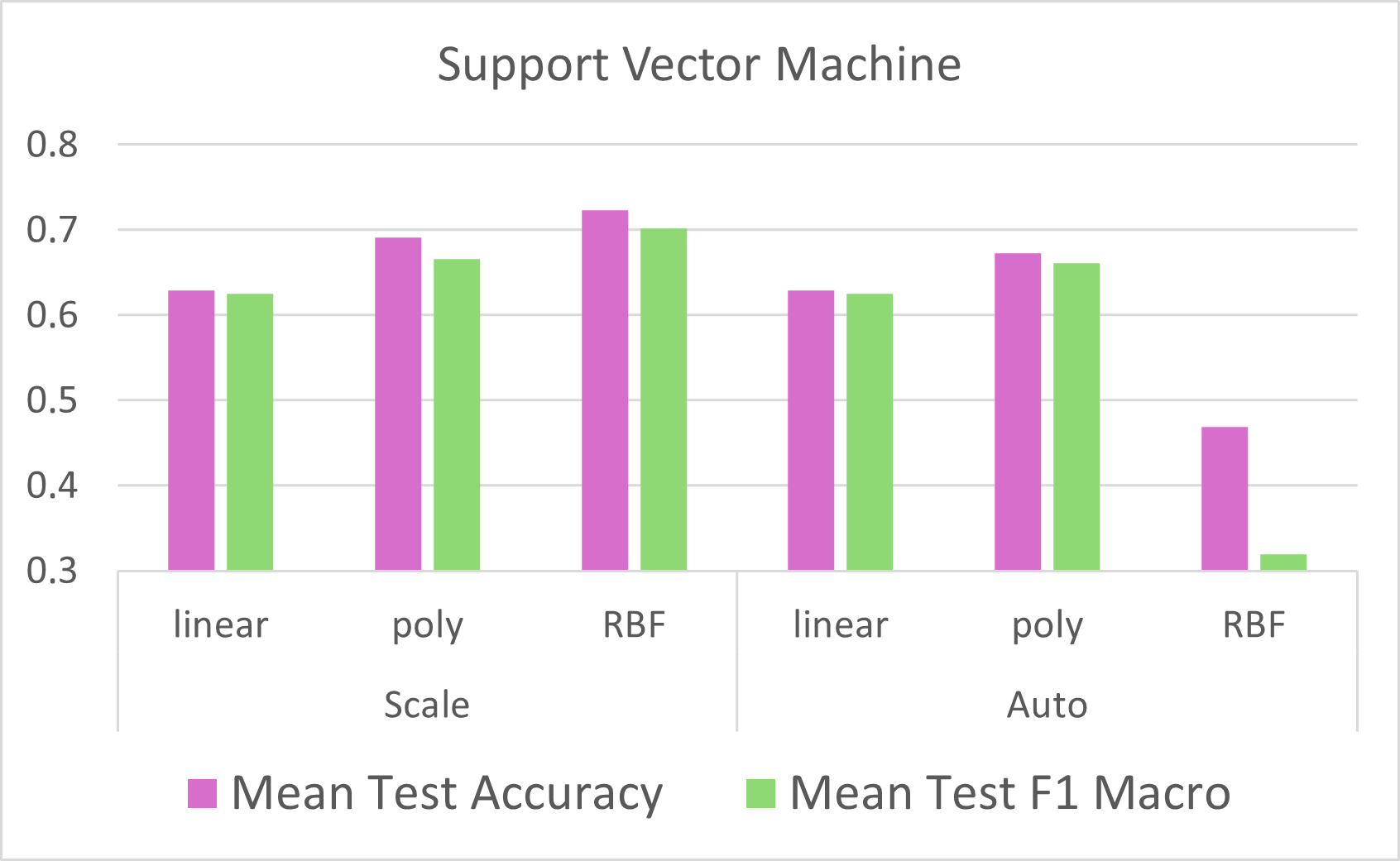}\label{subfig:SVM_ALL}}\quad
    \subfloat[][\centering]{\includegraphics[width=0.47\textwidth]{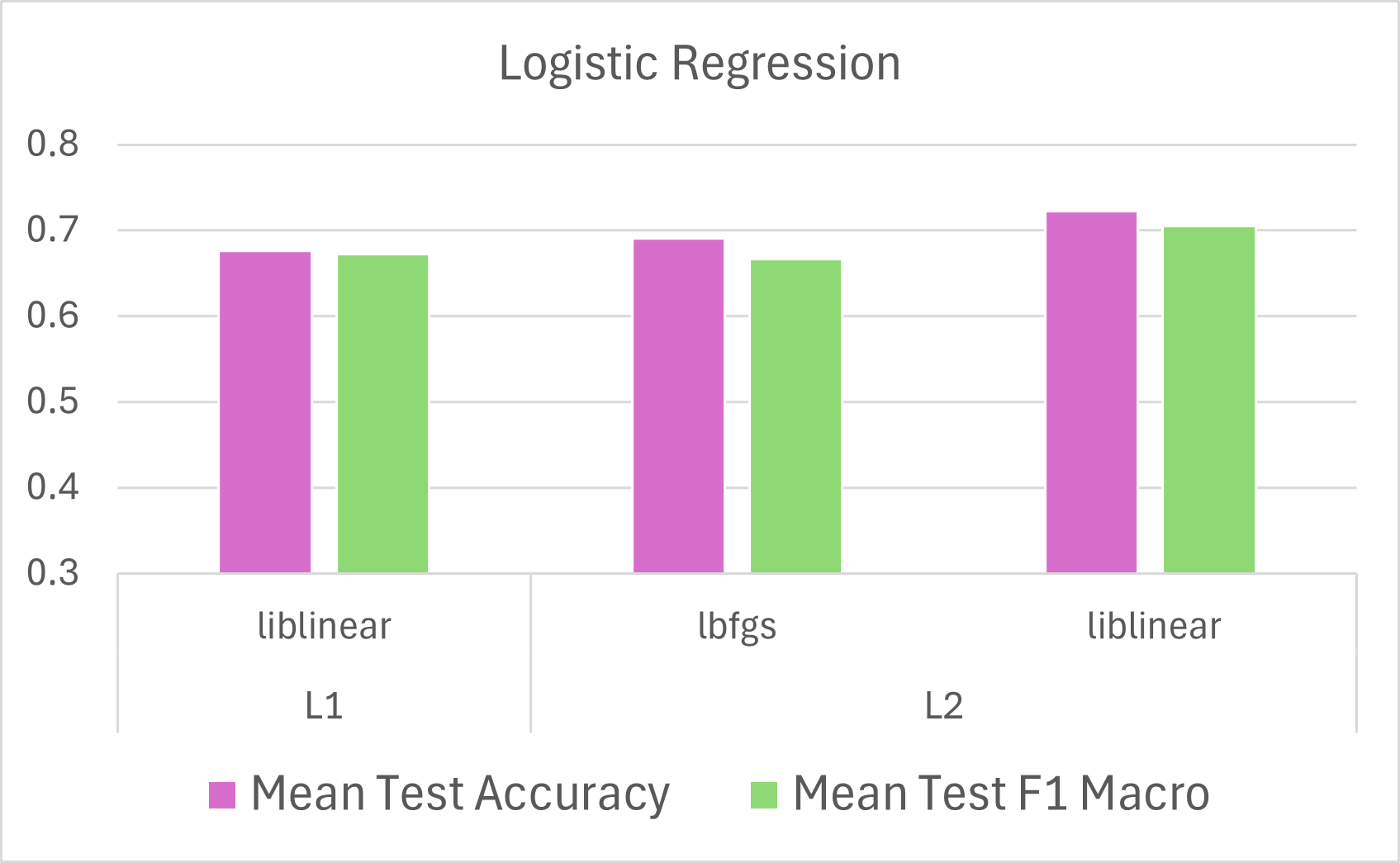}\label{subfig:LR_ALL}} \\
	\subfloat[][\centering]{\includegraphics[width=0.47\textwidth]{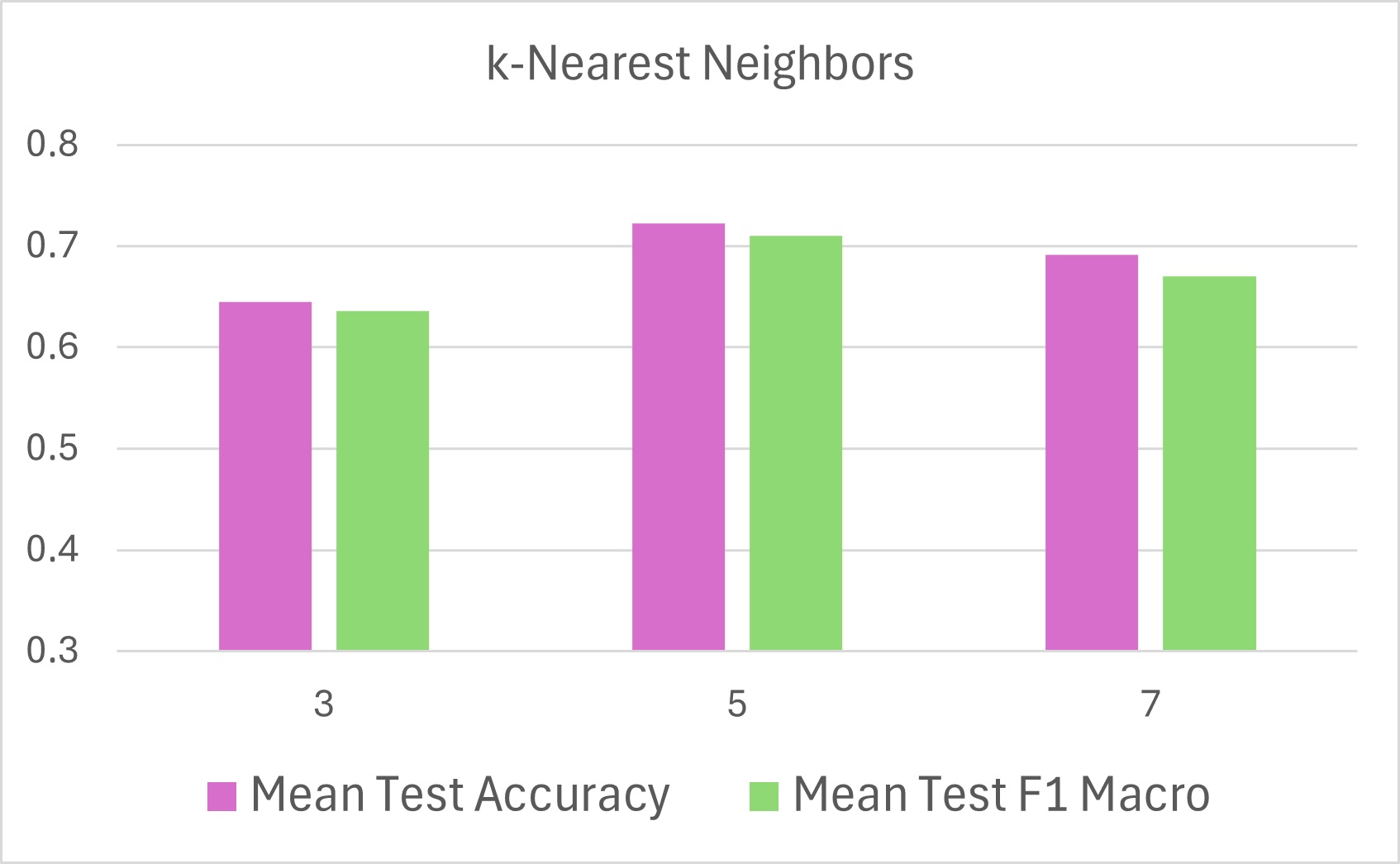}\label{subfig:KNN_ALL}}\quad
	\subfloat[][\centering]{\includegraphics[width=0.47\textwidth]{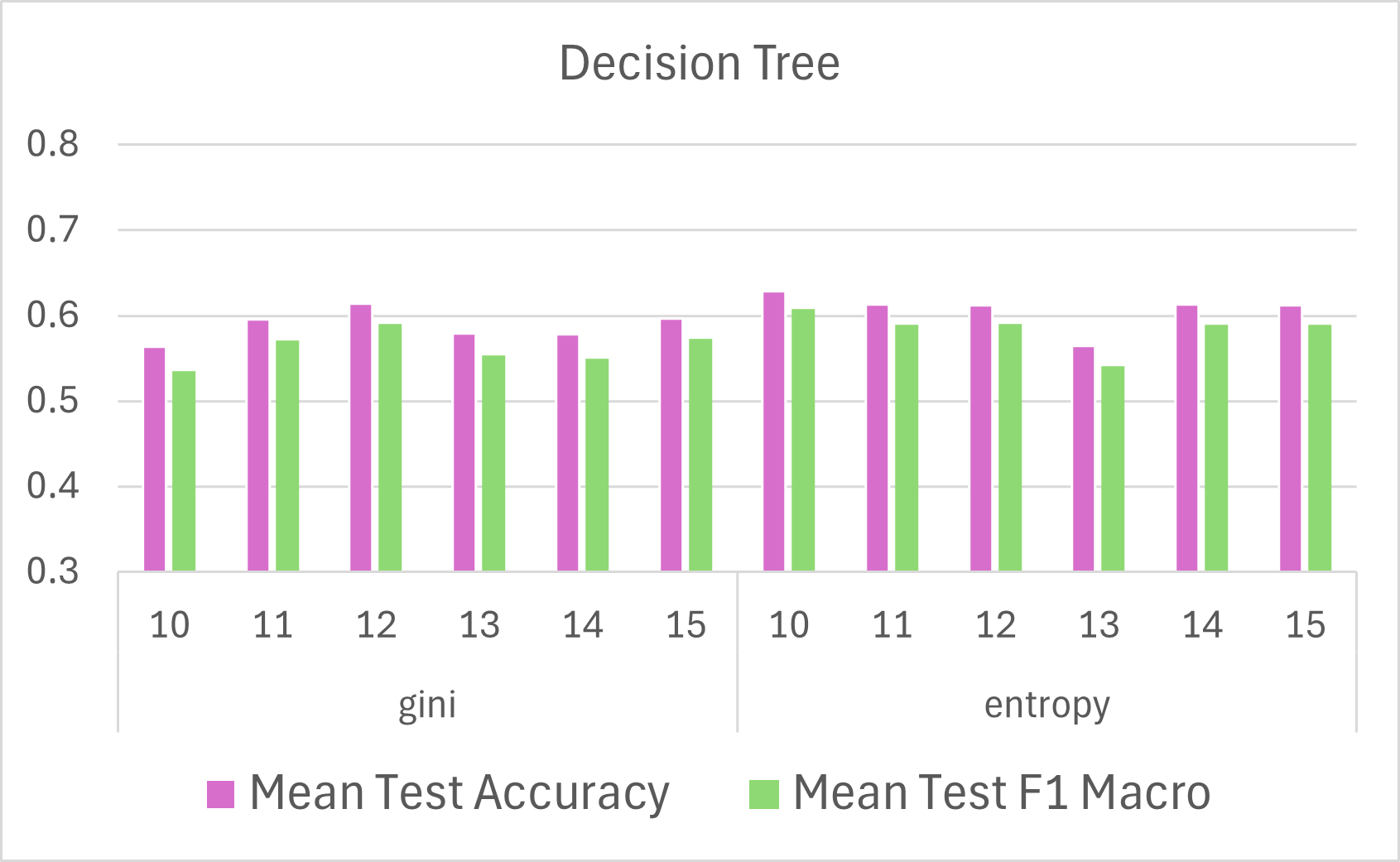}\label{subfig:DT_ALL}} \\
    \caption{{Performance} of the ML algorithms on the pooled group.  {(\textbf{a}) RF with gini criterion; (\textbf{b}) RF with entropy criterion; (\textbf{c}) SVM; (\textbf{d}) LR; (\textbf{e}) k-NN; (\textbf{f}) DT.}} 
    \label{fig:CombinedClass}
\end{figure}

When considering the pooled Italian and Spanish user group, the results achieved during the ML training validated both RF and SVM  as the top-performing algorithms. RF achieved a peak accuracy of 75.4\% and an F1-score of 73.3\%, Figure~\ref{fig:CombinedClass}a,b. This strong performance underscores RF's ability to generalize across different user demographics and potentially capture common patterns present in both groups. 
SVM also demonstrated robust performance in the pooled group, Figure~\ref{fig:CombinedClass}c. With the RBF kernel, SVM achieved an accuracy of 72.3\% and an F1-score of 70.2\%. This indicates that SVM remains a highly effective algorithm, although it was slightly outperformed by RF in this cross-language analysis. LR continued to provide consistent performance in the pooled group, with an accuracy and F1-score hovering around 70\%, Figure~\ref{fig:CombinedClass}d. This stability suggests that LR could reliably model the data, although it may not have achieved the highest possible predictive accuracy. k-NN was also competitive on the pooled group, achieving an accuracy of approximately 72\% with \mbox{five neighbors}, Figure~\ref{fig:CombinedClass}e. This performance level suggests that k-NN could effectively classify users in the pooled group, balancing between bias and variance. In contrast, the DT classifier remained the least effective algorithm for the pooled group, Figure~\ref{fig:CombinedClass}f, with accuracy scores generally in the low 60 s. This consistent under performance across all three groups reinforces the observation that DT may not be complex enough to capture the intricacies of user behavior compared to other algorithms.   \vspace{-5pt}

Again, the best configuration for performing the test was RF, but this time with an entropy criterion, a max depth of 15, and 20 estimators. The final scores for this configuration was 75.0\% for accuracy and 71.4\% for F1-score.

\section{Discussion}

This study explored the use of VR and ML to identify dyslexia in Italian and Spanish university students. The research encompassed two main areas: (i) analyzing differences in VR-derived behavioral data between students with and without SLDs, and (ii) evaluating the performance of ML algorithms in classifying dyslexia based on these data.

The statistical analysis revealed a pattern that partially supported the study's hypothesis. In the Italian sample, the \emph{t}-tests indicated significant differences in the time taken to complete the SR and RSES tasks (\emph{p} {< 0.001} and \emph{p} = $0.003$, respectively). This suggests that students with SLDs exhibited different task completion speeds compared to the control group. Importantly, the MW tests showed no significant differences in SR errors (\emph{p} = $0.584$) or RSES scores (\emph{p} = $0.531$), indicating that task accuracy and self-esteem levels did not significantly differ between the groups. These findings align with the expectation that dyslexia primarily manifests as a difference in processing speed, rather than accuracy or self-esteem. In contrast, the Spanish sample showed no statistical differences between groups for task completion time (\emph{t}-tests) or SR errors/RSES scores (MW tests). This suggests that, within the Spanish sample, the VR-derived data did not reveal significant differences between students with and without SLDs, in terms of speed, accuracy, or self-esteem. When the Italian and Spanish samples were combined, the results mirrored the Italian sample findings: significant differences in task completion time (\emph{p} $<0.001$ and \emph{p} = $0.005$) but no significant differences in SR errors or RSES scores (\emph{p} = $0.952$ and \emph{p} = $0.853$).  
Thus, the SR and RSES analyses statistically confirmed in two out of three cases how reading is a more time-consuming activity for people with SLDs~\cite{casini2018s,snowling2020defining};  Regarding the Spanish group, who did not pass the statistical analysis, but whose values remained close to the threshold limit, we can assume that the sample could be too meager to highlight this evidence. Staying on SR, the statistical score analysis of all three cases confirmed no difference between the two groups and that the results of individuals with SLDs can be the same as those of unaffected individuals, as long as it is supported with appropriate time and tools~\cite{moojen2020adults,van2019protective}. Finally, an interesting result lies in the self-esteem values noted during this study, where the MW test confirmed no statistical evidence among SLD and CG; in addition, while, as expected, the SLD group reported the lowest levels of self-esteem, they still achieved similar results to the CGs, who never managed to exceed the threshold value of 25 on the RSES scale that limited the low self-esteem area. This analysis confirms, as mentioned in the introduction, both that people with SLDs usually have low self-esteem~\cite{conley2007general,parshurami2015study,shah2019relationship,burden2005factors}, but also that this does not differ much from that of similarly aged peers without any disorder~\cite{pathrikar2016study}. Since the study was aimed at college students, the probable cause of these low values should be sought among academic challenges and social dynamics~\cite{arsandaux2023self,yang2023relationship}. 

The ML analysis demonstrated the feasibility of using VR-derived data to automatically identify dyslexia, but also revealed interesting differences in algorithm performance across the groups. On the test set, the classification accuracy was highest for the Italian group (87.5\% accuracy, 85.7\% F1-score), indicating the strong ability of the model to generalize to new data within the population. SVM is known for its effectiveness in high-dimensional spaces and its ability to capture complex non-linear relationships~\cite{rajeswari2024role,roy2023support}, which might explain its strong performance with the Italian data. The pooled group also showed promising results (75.0\% accuracy, 71.4\% F1-score), suggesting that the model could capture some cross-linguistic features relevant to dyslexia. RF, an algorithm robust to overfitting and capable of handling complex feature interactions~\cite{breiman2001random}, may have performed well in the combined group due to its ability to generalize across the increased variability introduced by combining the Italian and Spanish data~\cite{lovric2021machine}. However, the Spanish group showed a lower classification accuracy (66.6\% for both accuracy and F1-score), which might indicate that the model struggled to generalize to the specific characteristics of this group. This could have been due to differences in the nature of the VR-derived data itself, such as the feature distributions or the strength of the relationship between VR measures and dyslexia manifestation in Spanish speakers. The latter, in particular, emphasizes again how each language, as all orthographic systems, has its own challenges~\cite{moore2023cross}. Simpler algorithms like Logistic Regression and Decision Trees, while useful in other contexts, may have been less effective due to their limitations in capturing non-linear relationships and handling complex \mbox{feature interactions}.%

The cross-linguistic approach is a notable strength of this study. The successful application of ML algorithms, particularly on the Italian and entire datasets, suggests that the VR-based assessment captured some language-invariant features relevant to dyslexia. However, the lower performance on the Spanish group indicates that further investigation into language-specific influences is warranted.
\subsection{Limitations}
{The power analysis indicated that the design was highly sensitive to the large speed effects detected in the Italian and the pooled sample, but only moderately powered for SR time and under-powered for RSES time on the Spanish cohort, reinforcing the need for larger, age-diverse samples to confirm cross-linguistic generalization.}
{Regarding algorithmic fairness, no explicit fairness metrics, such as demographic parity or equal opportunity, were computed. However, the dataset was carefully balanced in terms of group composition (equal numbers of participants with and without dyslexia), as reported in Table~\ref{tab:demo}, as well as across key demographic variables such as age, language, and gender. This balanced design aimed to mitigate potential sources of bias during model training and evaluation. However, given the relatively small sample size, we avoided further subdividing the dataset to compute fairness-related metrics, as this would reduce the statistical power and compromise the reliability of the results. This limitation will be addressed in future studies involving larger and more diverse populations, where dedicated fairness assessments could be incorporated to ensure equitable performance across subgroups. Additionally, the lower performance observed in the Spanish dataset may have been due to several factors. The statistical differences between groups were weaker compared to the Italian sample, suggesting that the selected features may capture dyslexia-related patterns less effectively in Spanish. Although Spanish and Italian are both transparent languages, linguistic and cultural differences may affect how dyslexia manifests and how participants interacted with the VR tasks, demonstrating why a cross-linguistic approach is fundamental.}

\subsection{Future Works}
{Since the present study was conceived as an exploratory proof of concept, the VR protocol was not administered concurrently with the reference battery; a follow-up validation that presents equivalent {BDA 16-30}  passages in VR has already been planned.}
{Additionally, the BDA 16-30 manual is intended for a single administration, mainly to avoid passage familiarity, and test--retest coefficients are not provided in the normative tables, and the present study followed the same approach. Therefore, future work will develop parallel passages to estimate reliability without learning effects.
Regarding subsamples, the post hoc power of the Spanish subsample for SR-time was 0.52 and fell below 0.30 for the RSES-time; and medium cross-linguistic effects could therefore have escaped detection. A follow-up study will recruit a larger number of participants per language and will incorporate language as an explicit covariate in the classification model to test orthographic-transparency hypotheses with adequate statistical sensitivity.
Finally, although neither the Rosenberg self-esteem score nor the SR comprehension score showed a significant group difference, both were retained in this exploratory model because small distributional shifts, especially when combined with timing features, may still aid classification. Future studies will benchmark models with and without these accuracy measures, and will remove them if they add no \mbox{practical value}.}
\section{Conclusions}

This study provides valuable evidence for the potential of VR and ML in dyslexia assessment in Italian and Spanish university students. The machine learning results on the test set demonstrated the promise of these techniques for classifying dyslexia, with a particularly strong performance on the Italian group. The analysis of group differences largely supported the hypothesis that dyslexia is characterized by differences in task completion speed, with less impact on task accuracy and self-esteem. However, the difference in results between the Italian and Spanish samples and the varying performance of the ML models across groups suggest that every language needs a dedicated assessment, in order to select an appropriate algorithm. Future research should aim to expand the sample size, include participants from diverse age groups, and explore additional ML techniques. A key focus should be to further investigate the language-specific factors that may influence both VR-derived behavioral data and the performance of machine learning models.



\vspace{6pt} 


\authorcontributions{{Conceptualization},  M.M., G.M., J.M.A.-L., E.Y.-B., G.C., A.Z. and J.T.; Methodology, M.M., G.M., J.M.A.-L., A.Z. and J.T.; Resources, E.Y.-B. and G.C.; Data curation, M.M., G.M., J.M.A.-L., A.Z. and J.T.; Writing---original draft preparation, M.M., G.M., J.M.A.-L., A.Z. and J.T.; Writing---review and editing, M.M., G.M., J.M.A.-L., A.Z., E.Y.-B., G.C. and J.T.; Supervision, J.T., G.C. and E.Y.-B. All authors have read and agreed to the published version of the manuscript.}

\funding{{This research received no external funding}.}

\institutionalreview{Ethical review and approval were waived for this study, since no experimentation was carried out on human beings. The involvement of humans was limited to the completion of Virtual Reality tests.}

\informedconsent{{Informed consent was obtained from all subjects involved in \mbox{the study}.} 
}

\dataavailability{{The raw data supporting the conclusions of this article will be made available by the authors on
request.}
}

\acknowledgments{{José Manuel Alcalde-Llergo} enrolled in the National PhD in Artificial Intelligence, XXXVIII cycle, course on Health and Life Sciences, organized by Università Campus Bio-Medico di Roma. He is also pursuing his doctorate with co-supervision at the Universidad de Córdoba (Spain), enrolled in its PhD program in Computation, Energy and Plasmas.}

\conflictsofinterest{The authors declare no conflicts of interest.}

\begin{adjustwidth}{-\extralength}{0cm}
\reftitle{References}


\PublishersNote{}
\end{adjustwidth}

\begin{thebibliography}{999}

\bibitem[Organization()]{who}
World Health Organization. \newblock 6A03-Developmental Learning Disorder.
\newblock {Available online:}  \url{https://icd.who.int/browse/2024-01/mms/en\#2099676649} (accessed on 22 \mbox{April 2025}).

\bibitem[Benedetti et~al.(2022)Benedetti, Barone, Panetti, Taborri, Urbani,
Zingoni, and Calabrò]{Ilaria2022}
Benedetti, I.; Barone, M.; Panetti, V.; Taborri, J.; Urbani, T.; Zingoni, A.;
Calabrò, G.
\newblock Clustering analysis of factors affecting academic career of
university students with dyslexia in Italy.
\newblock {\em Sci. Rep.} {\bf 2022}, {\em 12},~9010. [\href{http://doi.org/10.1038/s41598-022-12985-w}{CrossRef}]

\bibitem[Moore et~al.(2023)Moore, Lai, Quinonez-Beltran, Wijekumar, and
Joshi]{moore2023cross}
Moore, K.A.; Lai, J.; Quinonez-Beltran, J.F.; Wijekumar, K.; Joshi, R.M.
\newblock A cross-orthographic view of dyslexia identification.
\newblock {\em J.~Cult. Cogn. Sci.} {\bf 2023}, {\em
7},~197--217. [\href{http://dx.doi.org/10.1007/s41809-023-00128-0}{CrossRef}]

\bibitem[Hengeveld and Leufkens(2018)]{hengeveld2018transparent}
Hengeveld, K.; Leufkens, S.
\newblock Transparent and non\-transparent languages.
\newblock {\em Folia Linguist.} {\bf 2018}, {\em 52},~139--175. [\href{http://dx.doi.org/10.1515/flin-2018-0003}{CrossRef}]

\bibitem[Cossu et~al.(1999)]{cossu1999acquisition}
Cossu, G.
\newblock The acquisition of Italian orthography. In
\newblock {\em Learning to Read and Write: A Cross-Linguistic Perspective}; {Cambridge University Press: Cambridge, UK,}  {1999}; Volume {2}, pp.~10--33.

\bibitem[Li et~al.(2018)Li, Han, Wang, Sun, and Cheng]{li2018social}
Li, J.; Han, X.; Wang, W.; Sun, G.; Cheng, Z.
\newblock How social support influences university students' academic
achievement and emotional exhaustion: The mediating role of self-esteem.
\newblock {\em Learn. Individ. Differ.} {\bf 2018}, {\em
61},~120--126. [\href{http://dx.doi.org/10.1016/j.lindif.2017.11.016}{CrossRef}]

\bibitem[Kermode and MacLean(2001)]{kermode2001study}
Kermode, S.; MacLean, D.
\newblock A study of the relationship between quality of life, self-esteem and
health.
\newblock {\em Aust. J. Adv. Nurs.} {\bf 2001}, {\em
19},~33--40. [\href{http://www.ncbi.nlm.nih.gov/pubmed/11845707}{PubMed}]

\bibitem[Donnellan et~al.(2015)Donnellan, Trzesniewski, and
Robins]{donnellan2015measures}
Donnellan, M.B.; Trzesniewski, K.H.; Robins, R.W.
\newblock Measures of self-esteem. In {\em Measures of Personality and Social
Psychological Constructs}; Elsevier:  {Amsterdam, The Netherlands,} 2015; pp. 131--157.


\bibitem[Rosenberg(1965)]{rosenberg1965rosenberg}
Rosenberg, M.
\newblock Rosenberg self-esteem scale.
\newblock {\em J. Relig. Health} { {\bf 1965}}. [\href{http://dx.doi.org/10.1037/t01038-000}{CrossRef}]


\bibitem[Tinakon and Nahathai(2012)]{tinakon2012comparison}
Tinakon, W.; Nahathai, W.
\newblock A comparison of reliability and construct validity between the
original and revised versions of the Rosenberg Self-Esteem Scale.
\newblock {\em Psychiatry Investig.} {\bf 2012}, {\em 9},~54. [\href{http://dx.doi.org/10.4306/pi.2012.9.1.54}{CrossRef}] [\href{http://www.ncbi.nlm.nih.gov/pubmed/22396685}{PubMed}]

\bibitem[Alesi et~al.(2012)Alesi, Rappo, and Pepi]{alesi2012self}
Alesi, M.; Rappo, G.; Pepi, A.
\newblock Self-esteem at school and self-handicapping in childhood: Comparison
of groups with learning disabilities.
\newblock {\em Psychol. Rep.} {\bf 2012}, {\em 111},~952--962. [\href{http://dx.doi.org/10.2466/15.10.PR0.111.6.952-962}{CrossRef}]

\bibitem[Conley et~al.(2007)Conley, Ghavami, VonOhlen, and
Foulkes]{conley2007general}
Conley, T.D.; Ghavami, N.; VonOhlen, J.; Foulkes, P.
\newblock General and domain-specific self-esteem among regular education and
special education students.
\newblock {\em J. Appl. Soc. Psychol.} {\bf 2007}, {\em
37},~775--789. [\href{http://dx.doi.org/10.1111/j.1559-1816.2007.00185.x}{CrossRef}]

\bibitem[Parshurami(2015)]{parshurami2015study}
Parshurami, A.
\newblock A study on self-esteem and adjustment in children with learning
disability.
\newblock {\em Indian J. Ment. Health} {\bf 2015}, {\em 2},~306--311. [\href{http://dx.doi.org/10.30877/IJMH.3.3.2015.306-310}{CrossRef}]

\bibitem[Shah(2019)]{shah2019relationship}
Shah, P.
\newblock The Relationship between Anxiety, Depression and Self-esteem in
Adolescents with Learning Disability.
\newblock {\em Indian J. Ment. Health} {\bf 2019}, {\em 6},~368--376. [\href{http://dx.doi.org/10.30877/IJMH.6.4.2019.368-376}{CrossRef}]

\bibitem[Burden and Burdett(2005)]{burden2005factors}
Burden, R.; Burdett, J.
\newblock Factors associated with successful learning in pupils with dyslexia:
A motivational analysis.
\newblock {\em Br. J. Spec. Educ.} {\bf 2005}, {\em
32},~100--104. [\href{http://dx.doi.org/10.1111/j.0952-3383.2005.00378.x}{CrossRef}]

\bibitem[Pathrikar(2016)]{pathrikar2016study}
Pathrikar, K.R.
\newblock A study on perceived social support and self esteem in children with
and without learning disability.
\newblock {\em Indian J. Ment. Health} {\bf 2016}, {\em{3},~271--277. [\href{http://dx.doi.org/10.30877/IJMH.3.3.2016.271-277}{CrossRef}]}


\bibitem[Catts and Petscher(2018)]{catts2018early}
Catts, H.; Petscher, Y.
\newblock Early identification of dyslexia: Current advancements and future
directions.
\newblock {\em Perspect. Lang. Lit.} {\bf 2018}, {\em
44},~33--36.

\bibitem[Zingoni et~al.(2021)Zingoni, Taborri, Panetti, Bonechi,
Aparicio-Mart{\'\i}nez, Pinzi, and Calabr{\`o}]{zingoni2021investigating}
Zingoni, A.; Taborri, J.; Panetti, V.; Bonechi, S.; Aparicio-Mart{\'\i}nez, P.;
Pinzi, S.; Calabr{\`o}, G.
\newblock Investigating issues and needs of dyslexic students at university:
Proof of concept of an artificial intelligence and virtual reality-based
supporting platform and preliminary results.
\newblock {\em Appl. Sci.} {\bf 2021}, {\em 11},~4624. [\href{http://dx.doi.org/10.3390/app11104624}{CrossRef}]

\bibitem[Zingoni et~al.(2024)Zingoni, Taborri, and
Calabrò]{vrailexiapredictor1}
Zingoni, A.; Taborri, J.; Calabrò, G.
\newblock A machine learning-based classification model to support university
students with dyslexia with personalized tools and strategies.
\newblock {\em Sci. Rep.} {\bf 2024}, {\em 14}, 273. [\href{http://dx.doi.org/10.1038/s41598-023-50879-7}{CrossRef}]

\bibitem[Morciano et~al.(2024)Morciano, Llergo, Zingoni, Bol{\'\i}var, Taborri,
and Calabr{\`o}]{morciano2024use}
Morciano, G.; Llergo, J.M.A.; Zingoni, A.; Bol{\'\i}var, E.Y.; Taborri, J.;
Calabr{\`o}, G.
\newblock Use of recommendation models to provide support to dyslexic students.
\newblock {\em Expert Syst. Appl.} {\bf 2024}, {\em 249},~123738. [\href{http://dx.doi.org/10.1016/j.eswa.2024.123738}{CrossRef}]

\bibitem[van~den Boer et~al.(2022)van~den Boer, Bazen, and
de~Bree]{van2022same}
van~den Boer, M.; Bazen, L.; de~Bree, E.
\newblock The same yet different: Oral and silent reading in children and
adolescents with dyslexia.
\newblock {\em J. Psycholinguist. Res.} {\bf 2022}, {\em
51},~803--817. [\href{http://dx.doi.org/10.1007/s10936-022-09856-w}{CrossRef}]

\bibitem[Smyrnakis et~al.(2021)Smyrnakis, Andreadakis, Rina, Boufachrentin, and
Aslanides]{smyrnakis2021silent}
Smyrnakis, I.; Andreadakis, V.; Rina, A.; Boufachrentin, N.; Aslanides, I.M.
\newblock Silent versus reading out loud modes: An eye-tracking study.
\newblock {\em J. Eye Mov. Res.} {\bf 2021}, {\em
14},~10--16910. [\href{http://dx.doi.org/10.16910/jemr.14.2.1}{CrossRef}]

\bibitem[Gagliano et~al.(2015)Gagliano, Ciuffo, Ingrassia, Ghidoni, Angelini,
Benedetto, German{\`o}, and Stella]{gagliano2015silent}
Gagliano, A.; Ciuffo, M.; Ingrassia, M.; Ghidoni, E.; Angelini, D.; Benedetto,
L.; German{\`o}, E.; Stella, G.
\newblock Silent reading fluency: Implications for the assessment of adults
with developmental dyslexia.
\newblock {\em J. Clin. Exp. Neuropsychol.} {\bf
2015}, {\em 37},~972--980. [\href{http://dx.doi.org/10.1080/13803395.2015.1072498}{CrossRef}]

\bibitem[Hairrell et~al.(2010)Hairrell, Edmonds, Vaughn, and
Simmons]{hairrell2010independent}
Hairrell, A.; Edmonds, M.; Vaughn, S.; Simmons, D.
\newblock Independent silent reading for struggling readers: Pitfalls and
potential.
\newblock In {\em Revisiting Silent Reading: New Directions for Teachers and Researchers}; International Reading Association: Newark, DE, USA, 2010; pp. 275--289, ISBN 979-0-87207-833-8.

\bibitem[Grigorenko et~al.(2020)Grigorenko, Compton, Fuchs, Wagner, Willcutt,
and Fletcher]{grigorenko2020understanding}
Grigorenko, E.L.; Compton, D.L.; Fuchs, L.S.; Wagner, R.K.; Willcutt, E.G.;
Fletcher, J.M.
\newblock Understanding, educating, and supporting children with specific
learning disabilities: 50 years of science and practice.
\newblock {\em Am. Psychol.} {\bf 2020}, {\em 75},~37. [\href{http://dx.doi.org/10.1037/amp0000452}{CrossRef}]

\bibitem[Kourtesis et~al.(2021)Kourtesis, Collina, Doumas, and
MacPherson]{kourtesis2021validation}
Kourtesis, P.; Collina, S.; Doumas, L.A.; MacPherson, S.E.
\newblock Validation of the Virtual Reality Everyday Assessment Lab (VR-EAL):
An immersive virtual reality neuropsychological battery with enhanced
ecological validity.
\newblock {\em J. Int. Neuropsychol. Soc.} {\bf
2021}, {\em 27},~181--196. [\href{http://dx.doi.org/10.1017/S1355617720000764}{CrossRef}]

\bibitem[Servotte et~al.(2020)Servotte, Goosse, Campbell, Dardenne, Pilote,
Simoneau, Guillaume, Bragard, and Ghuysen]{servotte2020virtual}
Servotte, J.C.; Goosse, M.; Campbell, S.H.; Dardenne, N.; Pilote, B.; Simoneau,
I.L.; Guillaume, M.; Bragard, I.; Ghuysen, A.
\newblock Virtual reality experience: Immersion, sense of presence, and
cybersickness.
\newblock {\em Clin. Simul. Nurs.} {\bf 2020}, {\em 38},~35--43. [\href{http://dx.doi.org/10.1016/j.ecns.2019.09.006}{CrossRef}]

\bibitem[Lin et~al.(2024)Lin, Li, Yao, Yang, and Zhang]{lin2024impact}
Lin, X.P.; Li, B.B.; Yao, Z.N.; Yang, Z.; Zhang, M.
\newblock The impact of virtual reality on student engagement in the
classroom---A critical review of the literature.
\newblock {\em Front. Psychol.} {\bf 2024}, {\em 15},~1360574. [\href{http://dx.doi.org/10.3389/fpsyg.2024.1360574}{CrossRef}]

\bibitem[Drigas et~al.(2022)Drigas, Mitsea, and Skianis]{drigas2022virtual}
Drigas, A.; Mitsea, E.; Skianis, C.
\newblock Virtual reality and metacognition training techniques for learning
disabilities.
\newblock {\em Sustainability} {\bf 2022}, {\em 14},~10170. [\href{http://dx.doi.org/10.3390/su141610170}{CrossRef}]

\bibitem[Alcalde-Llergo et~al.(2023)Alcalde-Llergo, Yeguas-Bol{\'\i}var,
Aparicio-Mart{\'\i}nez, Zingoni, Taborri, and Pinzi]{alcalde2023vr}
Alcalde-Llergo, J.M.; Yeguas-Bol{\'\i}var, E.; Aparicio-Mart{\'\i}nez, P.;
Zingoni, A.; Taborri, J.; Pinzi, S.
\newblock A VR serious game to increase empathy towards students with
phonological dyslexia.
\newblock In Proceedings of the 2023 IEEE International Conference on Metrology
for eXtended Reality, Artificial Intelligence and Neural Engineering
(MetroXRAINE),  {Milano, Italy, 25--27 October 2023}; \mbox{pp. 184--188}.

\bibitem[Alcalde-Llergo et~al.(2025)Alcalde-Llergo, Aparicio-Martínez,
Zingoni, Pinzi, and Yeguas-Bolívar]{alcalde2025}
Alcalde-Llergo, J.M.; Aparicio-Martínez, P.; Zingoni, A.; Pinzi, S.;
Yeguas-Bolívar, E.
\newblock Fostering Inclusion: A Virtual Reality Experience to Raise Awareness
of Dyslexia-Related Barriers in University Settings.
\newblock {\em Electronics} {\bf 2025}, {\em 14}, 829. [\href{http://dx.doi.org/10.3390/electronics14050829}{CrossRef}]

\bibitem[Pijnenborg et~al.(2022)Pijnenborg, Nijman, and
Veling]{pijnenborg2022discovr}
Pijnenborg, G.; Nijman, S.; Veling, W.
\newblock DiscoVR: Results of a multicenter RCT on a social cognitive virtual
reality training to enhance social cognition in psychosis.
\newblock {\em Eur. Psychiatry} {\bf 2022}, {\em 65},~S119. [\href{http://dx.doi.org/10.1192/j.eurpsy.2022.330}{CrossRef}]

\bibitem[Maskati et~al.(2021)Maskati, Alkeraiem, Khalil, Baik, Aljuhani, and
Alsobhi]{maskati2021using}
Maskati, E.; Alkeraiem, F.; Khalil, N.; Baik, R.; Aljuhani, R.; Alsobhi, A.
\newblock Using virtual reality (VR) in teaching students with dyslexia.
\newblock {\em Int. J. Emerg. Technol. Learn.} {\bf 2021}, {\em 16},~291--305. [\href{http://dx.doi.org/10.3991/ijet.v16i09.19653}{CrossRef}]

\bibitem[Maresca et~al.(2022)Maresca, Leonardi, De~Cola, Giliberto, Di~Cara,
Corallo, Quartarone, and Pidal{\`a}]{maresca2022use}
Maresca, G.; Leonardi, S.; De~Cola, M.C.; Giliberto, S.; Di~Cara, M.; Corallo,
F.; Quartarone, A.; Pidal{\`a}, A.
\newblock Use of virtual reality in children with dyslexia.
\newblock {\em Children} {\bf 2022}, {\em 9},~1621. [\href{http://dx.doi.org/10.3390/children9111621}{CrossRef}]

\bibitem[Maresca et~al.(2024)Maresca, Corallo, De~Cola, Formica, Giliberto,
Rao, Crupi, Quartarone, and Pidal{\`a}]{maresca2024effectiveness}
Maresca, G.; Corallo, F.; De~Cola, M.C.; Formica, C.; Giliberto, S.; Rao, G.;
Crupi, M.F.; Quartarone, A.; Pidal{\`a}, A.
\newblock Effectiveness of the Use of Virtual Reality Rehabilitation in
Children with Dyslexia: Follow-Up after One Year.
\newblock {\em Brain Sci.} {\bf 2024}, {\em 14},~655. [\href{http://dx.doi.org/10.3390/brainsci14070655}{CrossRef}] [\href{http://www.ncbi.nlm.nih.gov/pubmed/39061396}{PubMed}]

\bibitem[Vaitheeshwari et~al.(2024)Vaitheeshwari, Chih-Hsuan, Chung, Yang, Yeh,
Wu, and Kumar]{vaitheeshwari2024dyslexia}
Vaitheeshwari, R.; Chih-Hsuan, C.; Chung, C.R.; Yang, H.Y.; Yeh, S.C.; Wu,
E.H.K.; Kumar, M.
\newblock Dyslexia Analysis and Diagnosis Based on Eye Movement.
\newblock {\em IEEE Trans. Neural Syst. Rehabil. Eng.} {\bf 2024}, \emph{32}, 4109--4119. [\href{http://dx.doi.org/10.1109/TNSRE.2024.3496087}{CrossRef}] [\href{http://www.ncbi.nlm.nih.gov/pubmed/39527420}{PubMed}]

\bibitem[Chalkiadakis et~al.(2024)Chalkiadakis, Seremetaki, Kanellou, Kallishi,
Morfopoulou, Moraitaki, and Mastrokoukou]{chalkiadakis2024impact}
Chalkiadakis, A.; Seremetaki, A.; Kanellou, A.; Kallishi, M.; Morfopoulou, A.;
Moraitaki, M.; Mastrokoukou, S.
\newblock Impact of artificial intelligence and virtual reality on educational
inclusion: A systematic review of technologies supporting students with
disabilities.
\newblock {\em Educ. Sci.} {\bf 2024}, {\em 14},~1223. [\href{http://dx.doi.org/10.3390/educsci14111223}{CrossRef}]

\bibitem[Ko et~al.(2020)Ko, Jang, Lee, Yun, Kim, et~al.]{ko2020effects}
Ko, J.; Jang, S.W.; Lee, H.T.; Yun, H.K.; Kim, Y.S.
\newblock Effects of virtual reality and non--virtual reality exercises on the
exercise capacity and concentration of users in a ski exergame: Comparative
study.
\newblock {\em JMIR Serious Games} {\bf 2020}, {\em 8},~e16693. [\href{http://dx.doi.org/10.2196/16693}{CrossRef}]

\bibitem[Wen et~al.(2024)Wen, Gupta, Sasikumar, Billinghurst, Wilmott, Skow,
Dey, and Nanayakkara]{wen2024vr}
Wen, E.; Gupta, C.; Sasikumar, P.; Billinghurst, M.; Wilmott, J.; Skow, E.;
Dey, A.; Nanayakkara, S.
\newblock VR.net: A real-world dataset for virtual reality motion sickness
research.
\newblock {\em IEEE Trans. Vis. Comput. Graph.} {\bf
2024}, \emph{30}, 2330--2336. [\href{http://dx.doi.org/10.1109/TVCG.2024.3372044}{CrossRef}]

\bibitem[Spitzley and Karduna(2019)]{spitzley2019feasibility}
Spitzley, K.A.; Karduna, A.R.
\newblock Feasibility of using a fully immersive virtual reality system for
kinematic data collection.
\newblock {\em J. Biomech.} {\bf 2019}, {\em 87},~172--176. [\href{http://dx.doi.org/10.1016/j.jbiomech.2019.02.015}{CrossRef}]

\bibitem[Khan et~al.(2018)Khan, Cheng, and Bee]{khan2018machine}
Khan, R.U.; Cheng, J.L.A.; Bee, O.Y.
\newblock Machine learning and Dyslexia: Diagnostic and classification system
(DCS) for kids with learning disabilities.
\newblock {\em Int. J. Eng. Technol.} {\bf 2018},
{\em 7},~97--100.

\bibitem[Tamboer et~al.(2016)Tamboer, Vorst, Ghebreab, and
Scholte]{tamboer2016machine}
Tamboer, P.; Vorst, H.; Ghebreab, S.; Scholte, H.
\newblock Machine learning and dyslexia: Classification of individual
structural neuro-imaging scans of students with and without dyslexia.
\newblock {\em Neuroimage Clin.} {\bf 2016}, {\em 11},~508--514. [\href{http://dx.doi.org/10.1016/j.nicl.2016.03.014}{CrossRef}]

\bibitem[P{\l}o{\'n}ski et~al.(2014)P{\l}o{\'n}ski, Gradkowski, Marchewka,
Jednor{\'o}g, and Bogorodzki]{plonski2014dealing}
P{\l}o{\'n}ski, P.; Gradkowski, W.; Marchewka, A.; Jednor{\'o}g, K.;
Bogorodzki, P.
\newblock Dealing with the heterogeneous multi-site neuroimaging data sets: A
discrimination study of children dyslexia.
\newblock In \emph{Proceedings of the Brain Informatics and Health: International
Conference, BIH 2014, Warsaw, Poland, 11--14 August 2014}; Proceedings;
Springer: {Cham, Switzerland}, 
2014; \mbox{pp. 471--480}.


\bibitem[P{\l}o{\'n}ski et~al.(2017)P{\l}o{\'n}ski, Gradkowski, Altarelli,
Monzalvo, van Ermingen-Marbach, Grande, Heim, Marchewka, Bogorodzki, Ramus,
et~al.]{plonski2017multi}
P{\l}o{\'n}ski, P.; Gradkowski, W.; Altarelli, I.; Monzalvo, K.; van
Ermingen-Marbach, M.; Grande, M.; Heim, S.; Marchewka, A.; Bogorodzki, P.;
Ramus, F.;  et~al.
\newblock Multi-parameter machine learning approach to the neuroanatomical
basis of developmental dyslexia.
\newblock {\em Hum. Brain Mapp.} {\bf 2017}, {\em 38},~900--908. [\href{http://dx.doi.org/10.1002/hbm.23426}{CrossRef}]

\bibitem[Zhang et~al.(2024)Zhang, Lin, Yang, Chen, Cheng, and
Cheng]{zhang2024sample}
Zhang, L.; Lin, Y.; Yang, X.; Chen, T.; Cheng, X.; Cheng, W.
\newblock From sample poverty to rich feature learning: A new metric learning
method for few-shot classification.
\newblock {\em IEEE Access} {\bf 2024}, \emph{12}, 124990--125002. [\href{http://dx.doi.org/10.1109/ACCESS.2024.3444483}{CrossRef}]

\bibitem[Yeguas-Bol{\'\i}var et~al.(2022)Yeguas-Bol{\'\i}var, Alcalde-Llergo,
Aparicio-Mart{\'\i}nez, Taborri, Zingoni, and Pinzi]{yeguas2022determining}
Yeguas-Bol{\'\i}var, E.; Alcalde-Llergo, J.M.; Aparicio-Mart{\'\i}nez, P.;
Taborri, J.; Zingoni, A.; Pinzi, S.
\newblock Determining the difficulties of students with dyslexia via virtual
reality and artificial intelligence: An exploratory analysis.
\newblock In Proceedings of the 2022 IEEE International Conference on Metrology
for Extended Reality, Artificial Intelligence and Neural Engineering
(MetroXRAINE), {Rome, Italy, 26--28 October 2022}; pp. 585--590.


\bibitem[Materazzini et~al.(2024)Materazzini, Morciano, Alcalde-Llergo,
Yeguas-Bolivar, Zingoni, and Taborri]{materazzini24}
Materazzini, M.; Morciano, G.; Alcalde-Llergo, J.M.; Yeguas-Bolivar, E.;
Zingoni, A.; Taborri, J.
\newblock VR-based Silent Reading and Rosenberg Tests: Machine-Learning
Approach to Identify Learning Disorders.
\newblock In Proceedings of the 2024 IEEE International Conference on Metrology
for eXtended Reality, Artificial Intelligence and Neural Engineering
(MetroXRAINE), {St Albans, UK, 21--23 October 2024}; pp. 541--546. [\href{http://dx.doi.org/10.1109/MetroXRAINE62247.2024.10797202}{CrossRef}]


\bibitem[Zingoni et~al.(2024)Zingoni, Morciano, Alcalde-Llergo, Taborri,
Yeguas-Bolivar, Aparicio-Martinez, Pinzi, and Calabro]{vrailexiasummary}
Zingoni, A.; Morciano, G.; Alcalde-Llergo, J.M.; Taborri, J.; Yeguas-Bolivar,
E.; Aparicio-Martinez, P.; Pinzi, S.; Calabro, G.
\newblock VRAIlexia project: Provide customized support to university students
with dyslexia using Artificial Intelligence and Virtual Reality.
\newblock In Proceedings of the 2024 IEEE International Conference on Metrology
for eXtended Reality, Artificial Intelligence and Neural Engineering
(MetroXRAINE), {St Albans, UK, 21--23 October 2024}; pp. 535--540. [\href{http://dx.doi.org/10.1109/MetroXRAINE62247.2024.10796149}{CrossRef}]


\bibitem[Santulli et~al.(2018)Santulli, Scagnelli, Ciuffo, Baradello,
et~al.]{santulli2018superreading}
Santulli, F.; Scagnelli, M.; Ciuffo, M.; Baradello, A.
\newblock SuperReading: Ulteriori prove di efficacia rilevate con i test di
valutazione per l'adulto.
\newblock {\em Dislessia} {\bf 2018}, {\em 15},~35--51.

\bibitem[Ciuffo et~al.(2018)Ciuffo, Angelini, Barletta~Rodolfi, Gagliano,
Ghidoni, and Stella]{bda}
Ciuffo, M.; Angelini, D.; Barletta~Rodolfi, C.; Gagliano, A.; Ghidoni, E.;
Stella, G.
\newblock {\em BDA 16-30 Batteria Dislessia Adulti}; Giunti Psychometrics
S.r.l.: {Florence, Italy},  2018.

\bibitem[vra()]{vrailexia}
VRAIlexia.
\newblock Available online: \url{https://vrailexia.eu/} (accessed on
22 April 2025).

\bibitem[Casini et~al.(2018)Casini, Pech-Georgel, and Ziegler]{casini2018s}
Casini, L.; Pech-Georgel, C.; Ziegler, J.C.
\newblock It's about time: Revisiting temporal processing deficits in dyslexia.
\newblock {\em Dev. Sci.} {\bf 2018}, {\em 21},~e12530. [\href{http://dx.doi.org/10.1111/desc.12530}{CrossRef}]

\bibitem[Snowling et~al.(2020)Snowling, Hulme, and
Nation]{snowling2020defining}
Snowling, M.J.; Hulme, C.; Nation, K.
\newblock Defining and understanding dyslexia: Past, present and future.
\newblock {\em Oxf. Rev. Educ.} {\bf 2020}, {\em 46},~501--513. [\href{http://dx.doi.org/10.1080/03054985.2020.1765756}{CrossRef}]

\bibitem[Moojen et~al.(2020)Moojen, Gon{\c{c}}alves, Bass{\^o}a, Navas, de~Jou,
and Miguel]{moojen2020adults}
Moojen, S.M.P.; Gon{\c{c}}alves, H.A.; Bass{\^o}a, A.; Navas, A.L.; de~Jou, G.;
Miguel, E.S.
\newblock Adults with dyslexia: How can they achieve academic success despite
impairments in basic reading and writing abilities? The role of text
structure sensitivity as a compensatory skill.
\newblock {\em Ann. Dyslexia} {\bf 2020}, {\em 70},~115--140. [\href{http://dx.doi.org/10.1007/s11881-020-00195-w}{CrossRef}] [\href{http://www.ncbi.nlm.nih.gov/pubmed/32221905}{PubMed}]

\bibitem[van Viersen et~al.(2019)van Viersen, de~Bree, and
de~Jong]{van2019protective}
van Viersen, S.; de~Bree, E.H.; de~Jong, P.F.
\newblock Protective factors and compensation in resolving dyslexia.
\newblock {\em Sci. Stud. Read.} {\bf 2019}, {\em 23},~461--477. [\href{http://dx.doi.org/10.1080/10888438.2019.1603543}{CrossRef}]

\bibitem[Arsandaux et~al.(2023)Arsandaux, Boujut, Salamon, Tzourio, and
Gal{\'e}ra]{arsandaux2023self}
Arsandaux, J.; Boujut, E.; Salamon, R.; Tzourio, C.; Gal{\'e}ra, C.
\newblock Self-esteem in male and female college students: Does
childhood/adolescence background matter more than young-adulthood conditions?
\newblock {\em Personal. Individ. Differ.} {\bf 2023}, {\em
206},~112117. [\href{http://dx.doi.org/10.1016/j.paid.2023.112117}{CrossRef}]

\bibitem[Yang et~al.(2023)Yang, Huang, Li, Gan, Lin, and
Liu]{yang2023relationship}
Yang, S.; Huang, P.; Li, B.; Gan, T.; Lin, W.; Liu, Y.
\newblock The relationship of negative life events, trait-anxiety and
depression among Chinese university students: A moderated effect of
self-esteem.
\newblock {\em J. Affect. Disord.} {\bf 2023}, {\em 339},~384--391. [\href{http://dx.doi.org/10.1016/j.jad.2023.07.010}{CrossRef}]

\bibitem[Rajeswari and Kumari(2024)]{rajeswari2024role}
Rajeswari, S.; Kumari, D.A.
\newblock Role of Environmental, Social and Governance on Firm Value using
Support Vector Machine.
\newblock In Proceedings of the 2024 2nd International Conference on Advances
in Computation, Communication and Information Technology (ICAICCIT), {Faridabad, India, 28--29 November 2024}; Volume~1, pp. 524--529.


\bibitem[Roy and Chakraborty(2023)]{roy2023support}
Roy, A.; Chakraborty, S.
\newblock Support vector machine in structural reliability analysis: A review.
\newblock {\em Reliab. Eng. Syst. Saf.} {\bf 2023}, {\em
233},~109126. [\href{http://dx.doi.org/10.1016/j.ress.2023.109126}{CrossRef}]

\bibitem[Breiman(2001)]{breiman2001random}
Breiman, L.
\newblock Random forests.
\newblock {\em Mach. Learn.} {\bf 2001}, {\em 45},~5--32. [\href{http://dx.doi.org/10.1023/A:1010933404324}{CrossRef}]

\bibitem[Lovri{\'c} et~al.(2021)Lovri{\'c}, Pavlovi{\'c}, {\v{Z}}uvela,
Spataru, Lu{\v{c}}i{\'c}, Kern, and Wong]{lovric2021machine}
Lovri{\'c}, M.; Pavlovi{\'c}, K.; {\v{Z}}uvela, P.; Spataru, A.;
Lu{\v{c}}i{\'c}, B.; Kern, R.; Wong, M.W.
\newblock Machine learning in prediction of intrinsic aqueous solubility of
drug-like compounds: Generalization, complexity, or predictive ability?
\newblock {\em J. Chemom.} {\bf 2021}, {\em 35},~e3349. [\href{http://dx.doi.org/10.1002/cem.3349}{CrossRef}]

\end{thebibliography}
\end{document}